\documentclass[11pt]{article} 

\usepackage[utf8]{inputenc} 
\usepackage{color,graphicx}
\usepackage{caption}

\usepackage{geometry} 
\usepackage{amsmath}

\usepackage{booktabs} 
\usepackage{array} 
\usepackage{paralist} 
\usepackage{verbatim} 
\usepackage{subfig} 

\usepackage{sectsty}
\allsectionsfont{\sffamily\mdseries\upshape} 

\usepackage[nottoc,notlof,notlot]{tocbibind} 
\usepackage[titles,subfigure]{tocloft} 

\graphicspath{{./FiguresetResultats/}}
\usepackage{authblk}

\usepackage{fancyhdr} 
\begin{document}
\title{Organization and Independence or Interdependence? Study of the Neurophysiological Dynamics of Syntactic and Semantic Processing}
\author[1,2]{Sabine Ploux\thanks{Corresponding author: \texttt{sabine.ploux@ehess.fr}}}
\affil[1]{Institut des sciences cognitives-Marc Jeannerod CNRS/University of Lyon, Lyon, France}
\affil[2]{CAMS EHESS-CNRS, Paris, France}
\author[1,3]{Viviane Déprez}
\affil[3]{Rutgers, The State University of New Jersey, United States}
\date{}
\maketitle
\begin{abstract}
In this article we present a  multivariate model for determining  the different syntactic, semantic, and form (surface-structure) processes underlying the comprehension of simple phrases. 
 This model is applied to EEG signals recorded during a reading task. 
 The results show a hierarchical precedence of the neurolinguistic processes :  form, then syntactic  and lastly semantic processes. 
 We also found (a) that verbs are at the heart of phrase syntax processing, (b) an interaction between syntactic movement within the phrase, and semantic processes derived from a person-centered reference frame. 
Eigenvectors of the multivariate model provide electrode-times profiles that separate the distinctive linguistic processes and/or highlight their interaction. 
The accordance of these findings with different linguistic theories are discussed. 

\end{abstract}
\section{Introduction}
In this study, we try to determine the different syntactic, semantic, and surface-structure processes underlying the comprehension of two types of simple French phrases, in order to uncover their underlying neurophysiological bases, shared organization, and independence from, or on the contrary, interference with, each other. The two types of phrases examined are "subject verb object" phrases (SVO) like "la fillette mange la pomme" (the girl eats the apple) and "object relative-pronoun verb subject" phrases (O(rp)VS) like "la pomme que mange la fillette" (the apple eaten by the girl\footnote{A word-to-word English translation of this second type of French phrase — where the French relative pronoun "que" (that) refers to the object of the subordinate clause — gives the erroneous meaning "the apple that eats the girl". The correct meaning and can only be rendered in English by changing the word order ("the apple that the girl eats") or using the passive voice (the apple eaten by the girl). We use the passive voice translation in this study because it retains the O(rp)VS syntax tested here.}). To directly study the hierarchy of the underlying processes in terms of their linguistic functions and structures, temporal spans, interaction and/or independence, we applied a factor correspondence analysis (FCA) to the EEG signals obtained as the phrases were being read.

 Multivariate data analysis (including FCA) is being used increasingly often in neuroscience. When applied to fMRI data, it provides fine-grained information about the semantic processing of words and the brain networks involved; when applied to EEG data, it reveals the spatiotemporal networks (electrode-time units) of the lexicon's semantic structure. There are two good reasons for choosing this type of analysis to study the processes involved in the processing of phrases (as opposed to isolated words): (1) the ranking of the axes obtained from the analysis describes the hierarchy of the processes, and (2) the eigenvectors that define each of these axes give the neurophysiological profile (electrodes, temporal span) of the linguistic processes at play, thereby offering a spatiotemporal diagram of the processes analyzed. In sum, this type of analysis provides a conceptual framework whose organization is analogous to that of language, with its hierarchical structure, temporality, and interaction or independence of processes. We contend that because the dynamic, neurophysiological profile calculated from FCA eigenvectors is similar in format to that of language processes, it is useful for understanding language processes and also has the advantage of grasping those processes as a whole, including their complex organization and consistency. 

The specificity of our approach is detailed below from two angles: the study of the linguistic processes of syntax and semantics, and the fit between the study of those processes and the choice of neuroimaging methods and analyses.

\subsection{Separate approaches to syntax and semantics in neurolinguistics}
The faculty of language was first studied from the standpoint of behavioral deficits that develop following brain damage. This faculty — and more specifically, syntax and lexical semantics — was then studied in healthy subjects on the basis of linguistic theories. 
According to Chomsky, language is anchored in deep universal mechanisms (irrespective of the variability and manifestations of each particular language) that are directly imprinted in our genomic heritage; and the scientific goal of linguistics is to determine the universal rules and deep structures (competence) that enable the acquisition of a given language (performance). These universal rules of language have often been described in terms of three main modules: phonology (study of the structure of the phonic system of languages), syntax (study of word structures and combinations), and semantics (study of meaning). A large body of neurolinguistic research has been devoted to determining the neural bases of linguistic competence, using experimental protocols grounded in the original generative theory of language\cite{chomsky1957syntactic}, with these  distinctions and their modularity being posited as givens.This widespread approach attempts to understand the neural bases of the faculty of language, and consists of validating or invalidating the results or conclusions drawn from generative linguistic theories (\cite{zaccarella2017building, musso2003broca}). 

Accordingly, for syntax, many studies have tried to locate that cortical areas involved in the process that combines words into embedded syntactic structures in the form of trees, and to precisely determine the function of each of these brain areas (\cite{pattamadilok2016role,grodzinsky2000neurology,tyler2011left}).
	In parallel, research on semantics has attempted to determine the cerebral areas or networks involved in the semantic processing of words (\cite{huth2012continuous, ploux2012, huth2016natural}), independently of the argument structure and thus the syntactic structure induced by the words (verbs, for example \cite{Pulvermuller2005,hauk}).
	Lastly, studies whose protocols include both semantic and syntactic variables are aimed at detecting the neural networks selectively involved in the processing of these different types of variables (\cite{dapretto1999form,wehbe2014simultaneously}), based on the assumption that they are separate processes in the language. 
	
	However, some linguists who oppose these modularist postulates see the limitations of making a clear-cut distinction between syntax and semantics. Culioli said that "\emph{it is allowable ... to posit ... that at a very deep level (most likely prelexical), there exists a grammar of primitive relations where the distinction between syntax and semantic is meaningless}" (\cite{culioli1970considerations} ; our translation). And Rastier stated "\emph{that there exists a semantics of syntax, and even that syntax is semantics, from top to bottom}" (\cite{Rastier2005} ; our translation). 

In the present study, we take a neurophysiological view, in an attempt to distinguish between processes related to form (surface structure\footnote{Hereafter, we will use the term "form" to refer to the surface structure. In our stimuli, the form of a phrase corresponds to the position of the words in it.}), syntax , and semantics  during simple-phrase comprehension. This view allows us to look directly at a fine-grained decomposition of linguistic processes without distinguishing them in advance. We explain our approach — which exhibits good temporal resolution — on the basis of hypotheses drawn from the field of neurology.

\subsection{Fit between the structure of neurophysiological processes and the structure of linguistic processes }
Progress in neurolinguistics, made possible thanks to advances in neuroimaging, has been achieved in two main research domains: drawing up a map of brain areas and networks involved in language processing, and describing the temporality of those same processes. Each of these domains uses neuroimaging techniques: fMRI for brain mapping, EEG for temporality, and MEG in both cases since it advantageously offers temporal resolution and spatial resolution. In addition, the recent use of multivariate and/or deep-learning methods for processing neurophysiological signals \cite{Mitchell,huth2012continuous, ploux2012, huth2016natural} has opened up new avenues for obtaining more precise results and a better fit between language phenomena and the corresponding processing taking place in the brain. The description we give below of our approach will revolve around this fit between the direct study of language-specific phenomena and their underlying neural inscription. 

Progress in brain mapping has improved our understanding of the links between cortical areas or subareas, and language functions. However, new questions have emerged from this approach\footnote{One of the issues we might mention concerns the interpretation of brain maps, although we cannot address this question here because we used EEG and thus have no means of answering it. The brain-mapping research is aimed at finding the precise locations of linguistic phenomena. But the topology of the cerebral areas and the origin and reasons for that topology (like those of all other cognitive functions, for that matter), have not, to our knowledge, been sufficiently studied. Yet finding out about the origins and reasons behind the topology could supply information about the nature and differentiation of language functions: not only the nature and differences between the components of language but also their autonomy or integration into the set of all cognitive functions.}. The sole knowledge of a location (where?) does not give direct access to an understanding of the way in which the brain produces and processes linguistic units (how?). In other words, knowing where different linguistic processes take place does not offer any reasons for the specific structure of the complex, embedded, hierarchical\footnote{Hierarchical in the sense that the constituents of a phrase may be nested in other larger ones, or may even govern other constituents.} units of the language. As we know, the structure of language has the temporal axis as its communication medium. Imaging techniques that have a good temporal resolution, such as EEG and MEG, have contributed to advances in our understanding of the chronology of linguistic processing. By studying processing markers and their temporality, these techniques provide answers to the question of the neurophysiological anchoring of the components of language. For example, the pioneering work by Kutas  \cite{kutas1980reading} found evidence of the effects of a semantic distortion on wave N400 during sentence reading. Later work, most of which has been based on protocols involving a semantic or syntactic violation, has focused on supplying a functional interpretation (syntactic and/or semantic) of the different waves observed (principally waves ELAN, N400, and P600 \cite{frisch2004word}). These particular waves have become the principal object of investigations aimed not only at understanding the linguistic reasons behind their variations, but also at tracking the sequencing and automaticity of the syntactic and/or semantic processes associated with them, and at detecting their potential interactions as well\footnote{However, and as recalled above, these interactions have mainly been studied from the standpoint of cooperation and integration between the syntactic and semantic systems, defined separately (\cite{kim2011conflict,malaia2015neural}), so they are not necessarily seen as manifesting any kind of sharing of a common essence. Note that a recent study (\cite{van2016semantics}) obtained results supporting the idea that both syntax and semantics involve the activation of the sensorimotor brain system.}.

	Whatever the case may be, a process chronology that looks like the linear, surface order of sentences does not suffice to account for their complex, deep structure. In this study, in order to track the linguistic processes at play, their mutual organizations, and their temporality, we chose to analyze the data obtained from all of EEG electrodes. The use of the entire set of signals allowed in the data analysis offers a high degree of precision and more discrimination power than methods limited to certain regions, electrodes, or waves of interest. What's more, FCA eigenvectors describe an organization whose components can share intervals of time and are not necessarily sequential. Unlike other methods such as discriminant analysis, FCA is not supervised and does not prejudge groupings and differences between the experimental conditions. Lastly, in the present study, the stimulus phrases are totally understandable and did not contain any semantic or syntactic violations.

\section{Experiments}
\subsection{Experimental materials and task}
Phrases  were constructed from a list of words used in two earlier studies (\cite{ploux2012,plouxsubmitted}). The list contains 240 target words from eight
categories, four of which are biological (\emph{animals, people, fruits/vegetables, body parts}) and four of which are non-biological (\emph{clothing, tools, vehicles, household items}). The following features were controlled for the set of words as a whole, and for each of the eight categories: lexical frequency ($M(log10) = 0.9$, $SD = 0.9$), number of letters ($M = 6.4$, $SD = 1.6$), number of syllables ($M = 1.8$, $SD = 0.6$), number of phonemes ($M = 4.7$, $SD = 1.3$), grammatical gender (equal number of masculine as feminine nouns), and frequency of orthographic ($M = 2.9$, $SD = 3.6$) and phonological ($M = 8.0$, $SD = 8.3$) neighbors. The target words were put in subject position for half of the phrases, as in "le \textbf{bonnet} dissimule la chevelure" (\emph{the \textbf{hat} hides the hair}) and in object position for the other half, as in "la skieuse met le \textbf{bonnet}" (\emph{the skier puts on the \textbf{hat}}). In both cases, two phrases were constructed, one of the SVO type, as in "la skieuse met le \textbf{bonnet}" (\emph{the skier puts on the \textbf{hat}}) and one with the subject and verb occurring in the reverse order: O(rp)VS, as in "le \textbf{bonnet} que met la skieuse" (\emph{the \textbf{hat} put on by the skier}). The other words used in each phrase along with the target word (verb and subject or object) were selected from a large database of commonly associated words derived from novels and newspapers \cite{jibd1}. We made sure that the phrases did not contain any semantic violations. The full set of stimuli contained 420 standard-order phrases (240 x 2) and 420 reverse-order phrases (240 x 2).

	Having two word orders — standard order in SVO phrases and reverse order in O(rp)VS phrases — allowed us to separate a word's syntactic function from its location in the phrase\footnote{For the sake of readability, we will hereafter refer to these phrases in terms of their word order (standard vs. reverse), with "standard order" being the order of declarative sentences in French.}. Two experiments were conducted, one with standard-order phrases and one with reverse-order phrases. In both experiments, each word in the phrase was displayed alone in the center of the screen. Content words (the subject, verb, and object) were shown for 500 ms, function words (determiners and pronouns) for 110 ms. A period (".") was displayed for 500 ms at the end of each phrase, which was then followed by a black screen (530 ms). To keep the subject's attention, one out of twenty phrases (in random order) had a cross instead of a period at the end, followed by a word. The subject had to state whether the word in question was contained in the phrase just seen. To respond, the subject had to left click (vs. right click) on the mouse. A given subject saw only standard-order phrases (or reverse-order phrases). This avoided any effects of surprise or adjustment to the structure of the phrase. Both experiments began with a practice phase (10 phrases), followed by three blocks, each containing the whole set of phrases (standard or reverse) separated by a break. Within each block, the phrases were presented in random order.

\subsection{Participants}
There were 36 participants in all, 18 in the standard-order experiment (11 women, 7 men, mean age 23) and 18 in the reverse-order experiment (11 women, 7 men, mean age 22). They all reported having normal or corrected-to-normal vision. The study was approved by the CPP Sud-Est II Ethics Committee and all participants gave their written informed consent. 

\subsection{Signal acquisition} 
Presentation software ($^{\mbox{\scriptsize{ \texttrademark}}}$Neurobehavioral Systems) was used to program the experimental task. The EEG signals were recorded by an EGI recorder ($^{\mbox{\scriptsize{ \texttrademark}}}$Electrical Geodesics) connected to 128 electrodes placed on a cap helmet in which the impedances were kept under $50 k\Omega$. The sampling frequency was $1000 Hz$.

\section{Data analysis}
The EEG data were analyzed with Brainvision Analyser software ($^{\mbox{\scriptsize{ \texttrademark}}}$Brain Products). Phase-shift-free Butterworth filters were applied (high-pass: 0.1 Hz; low-pass: 100 Hz; notch: 50 Hz; slope: 48 dB/octave). 
Segmentation windows started $250 ms$ before the onset of the first word in each condition. Segments containing an amplitude greater than $\pm 100 mV$ were rejected. The mean of the segments was then calculated for each condition and each subject, along with the grand mean per condition for all subjects. A baseline correction was applied using a $[-250,0] ms$ window before stimulus onset. 

	For all tests presented here, the analysis method was FCA. Except for the overall analysis of the phrases (see Section 4 below), the FCA was applied to a matrix containing the conditions in the rows and the electrode-time units for a given time window in the columns. Each cell in the matrix contains the mean value of the signal ($\mu v$) for a given condition on a given electrode at a given time (defined at the precision level corresponding to the choice of signal segmentation). The calculations included (1) a map of the conditions showing their respective distances and proximities, (2) the profile of the associated eigenvector of each axis of the map, represented by the contribution to the electrode$-$time axis and by their topography for the contribution maxima. The profiles depict the spatiotemporal networks underlying the differentiation of the set of conditions analyzed. They rank the greatest differences between the conditions in decreasing order of importance, by axis number.

\section{Overall analysis of the phrases}
\begin{figure}
\hspace*{-1cm}
\includegraphics[width=16cm]{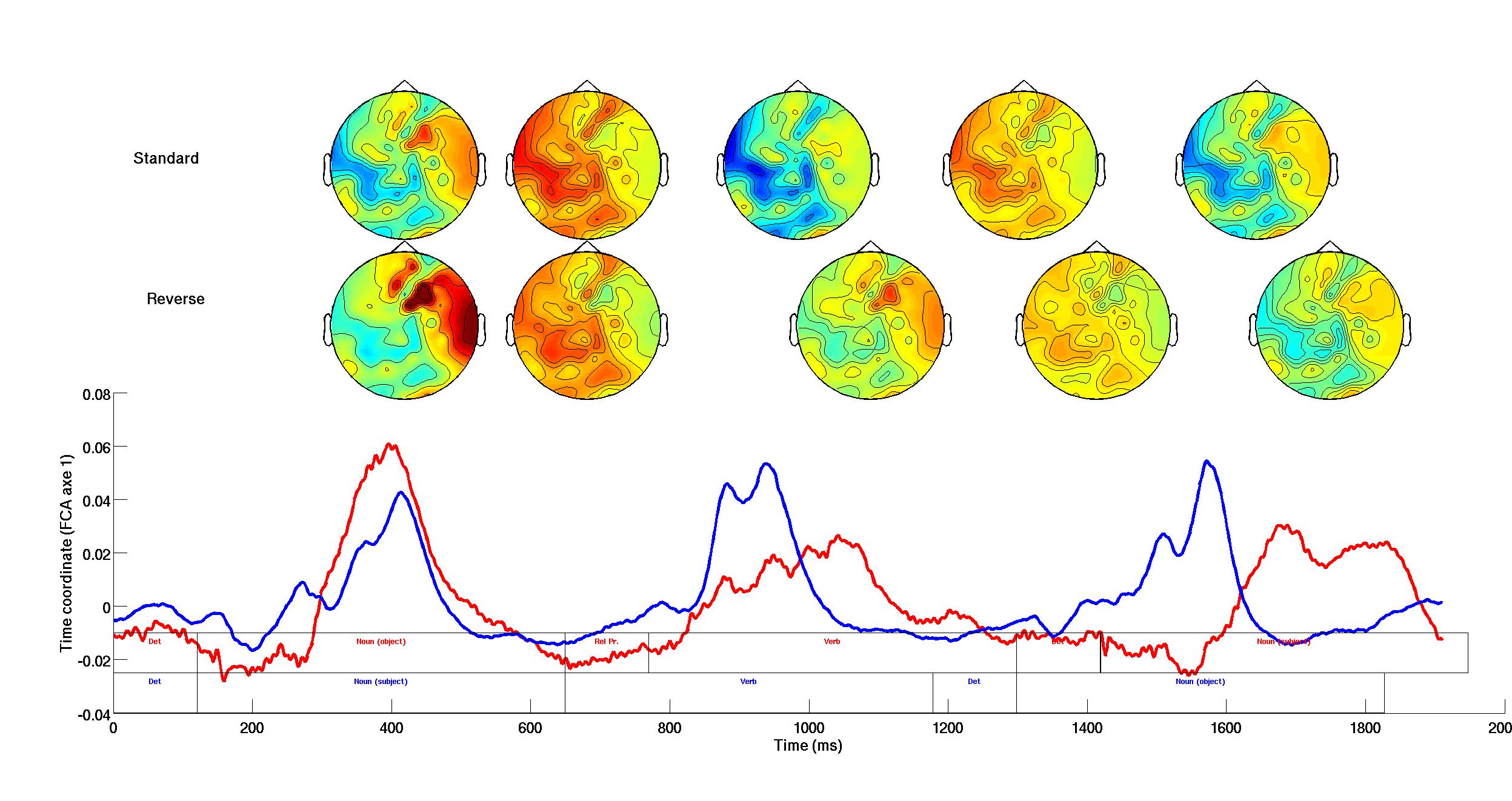}\\
\hspace*{-1cm}
\includegraphics[height=3cm, width=16cm]{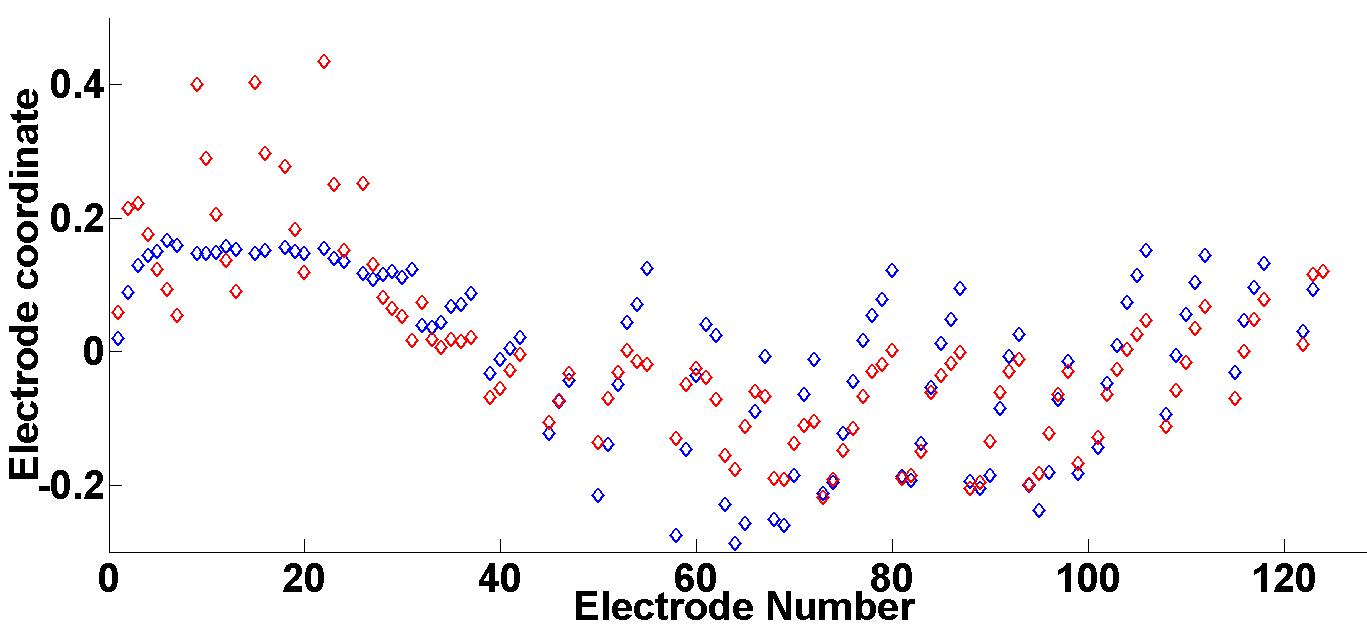}
\caption{Upper: Curves of the Axis-1 coordinates of the time variable in both FCAs (red: reverse order, blue: standard order). Above the curves, we can see the ERPs of the grand means in the standard-order and reverse-order conditions at the curves' maxima and minima, which correspond to the instants (detected by the FCA) where the difference between the signal on all electrodes and the signal over the whole time window under study (0-1850 ms) was the greatest. Axis 1 represents 42\% of the total inertia for standard phrases and 35\% for reverse phrases. Lower: Axis-1 coordinates of the electrodes in both FCAs (red: reverse order, blue: standard order). 
}
\label{phrases}
\end{figure}

In order to study the main processing components of the entire set of phrases, we ran two FCAs on the matrix of the ERP means of 102 electrodes, i.e., all electrodes except the peripheral ones (lower forehead, nose, cheeks, contour of the ears). The first FCA was applied to the standard-order phrases, the second to the reverse-order phrases. The principal axis in both FCAs pointed to the same processing and segmentation structure for content words (see Figure 1): the maximum values of the curve for each of the two types of phrases occurred 300 ms after the presentation of the content word; the minimum values occurred on each side of the content word ($-50$ to $100 ms$,  $450-600 ms$ from display onset). These maxima and minima correspond to the greatest contributions (in absolute value) of the electrodes. Figure  \ref{phrases} (lower) compares the electrode coordinates of the two types of phrases. One can see a great deal of similarity between the two profiles. The electrodes with the largest contributions are the mid-frontal, electrodes and (usually on the left) the frontal, temporal, and occipital electrodes. 

The fact that the waves of the two curves (Figure \ref{phrases} upper half) reveal analogous processing for the two types of phrases, centered at the instant located 300 ms after the onset of the content word, whose ending began while the word was still being displayed. This means that these waves were not determined solely by the presentation pace but also partially reflect the processes that define the phrase units. The second axis of both FCAs also pointed out a nearly periodic rhythm that aligns with the list of content words. Thus, the results of this analysis are more indicative of a type of processing of the succession and/or of the separation of content words. At this point, we have no evidence of structural differences within or between the two types of phrases.

	In this first analysis, whenever all lexical units in the phrase exhibited similar processing on the first and second axes, which together accounted for 58\% (standard phrases) and 59\% (reverse phrases) of the total inertia, we compared the different lexical units to each other to find functional distinctions. For this, we conducted a series of comparative analyses of content-word processing, starting with the entire set of words and then, step by step, taking each subset of words revealed by the previous analyses.

\section{Comparative study of the linguistic components of lexical-unit processing}
In order to study the linguistic components at play at during lexical-unit processing, we determined the grand mean of the 64 conditions, defined as follows: eight conditions (four biological categories and four non-biological categories) for grammatical subjects in phrase-initial position of standard-order phrases, eight conditions for grammatical objects in phrase-final position of standard-order phrases, eight conditions for grammatical objects in phrase-initial position of reverse-order phrases, eight conditions for grammatical subjects in phrase-final position of reverse-order phrases, sixteen conditions for verbs in standard-order phrases according to whether the verb followed or preceded a noun in one of the eight categories, and sixteen analogous conditions for verbs in reverse-order phrases. For each condition, the interval analyzed was $50 ms$ to $500 ms$ starting when the word was displayed. The baseline was calculated over the interval between $-250 ms$ and $0 ms$.

\subsection{Nouns and verbs: standard order and reverse order}
\begin{figure}
     \begin{center}
     \begin{tabular}{  p{5cm}  p{9cm} }
     \hline
\multicolumn{2}{l}{All content words}\\
\includegraphics[width=0.3\textwidth]{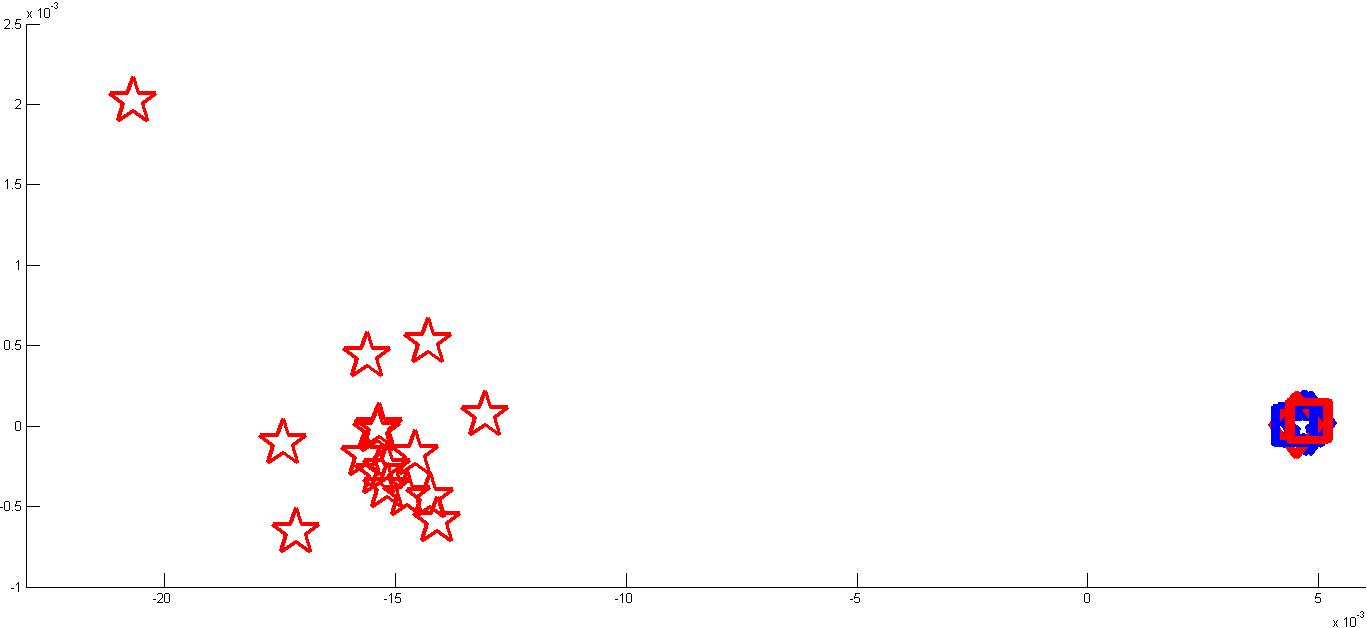}&\includegraphics[width=0.7\textwidth, height=20mm]{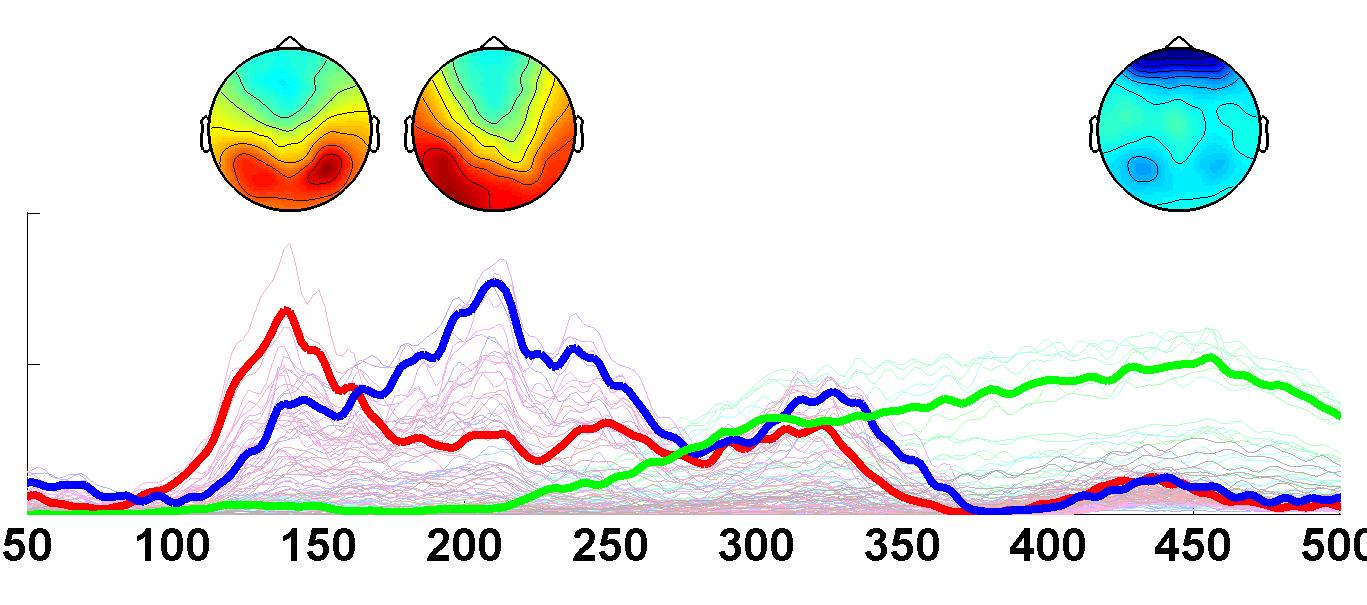}\\
\multicolumn{2}{l}{All nouns and standard-order verbs}\\
\includegraphics[width=0.3\textwidth]{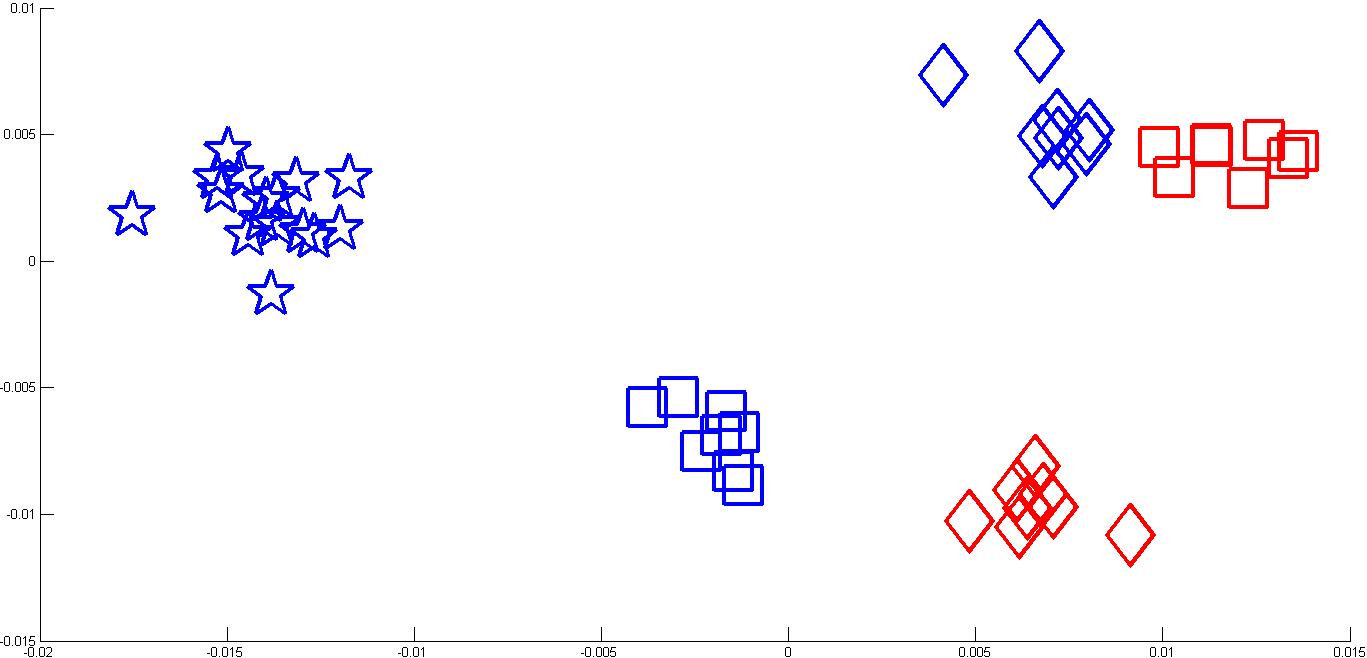}&\includegraphics[width=0.7\textwidth, height=20mm]{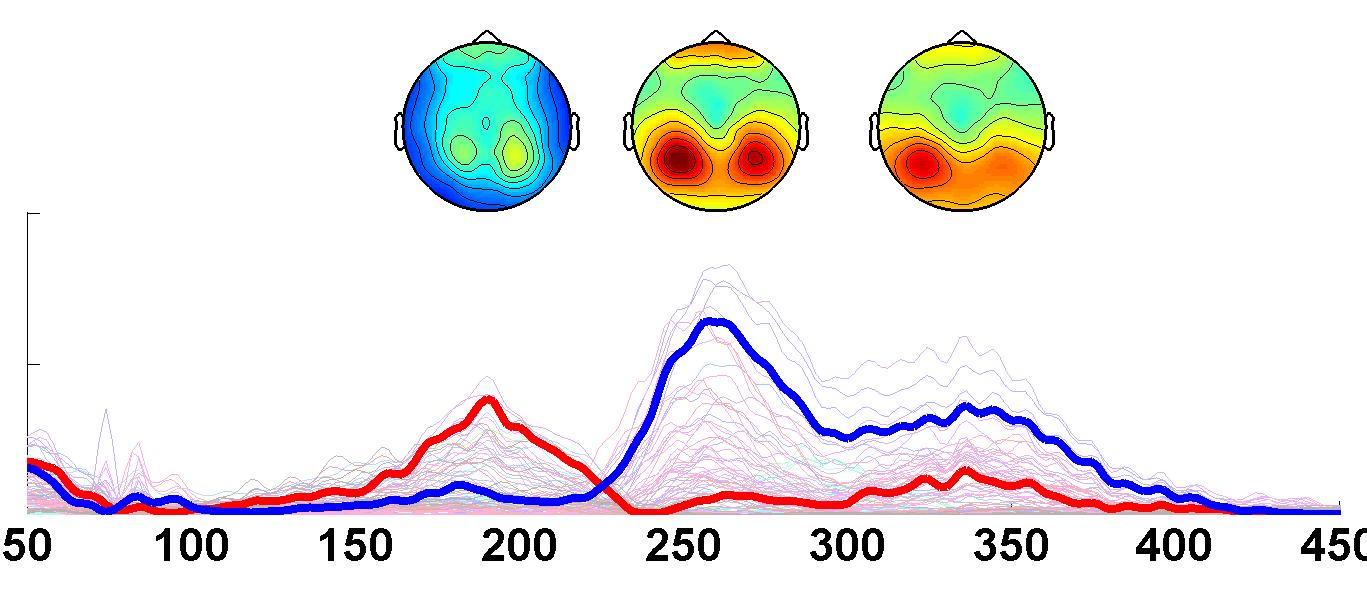}\\
\multicolumn{2}{l}{All verbs}\\
\includegraphics[width=0.3\textwidth, height=20mm]{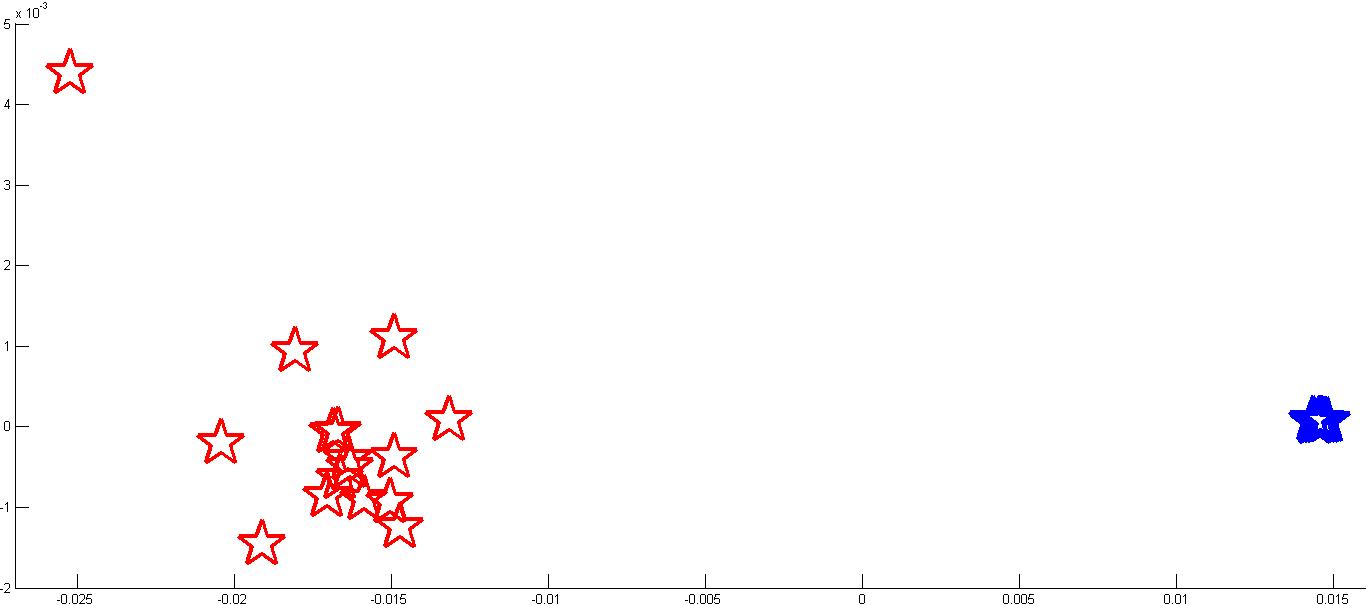}&\includegraphics[width=0.7\textwidth, height=20mm]{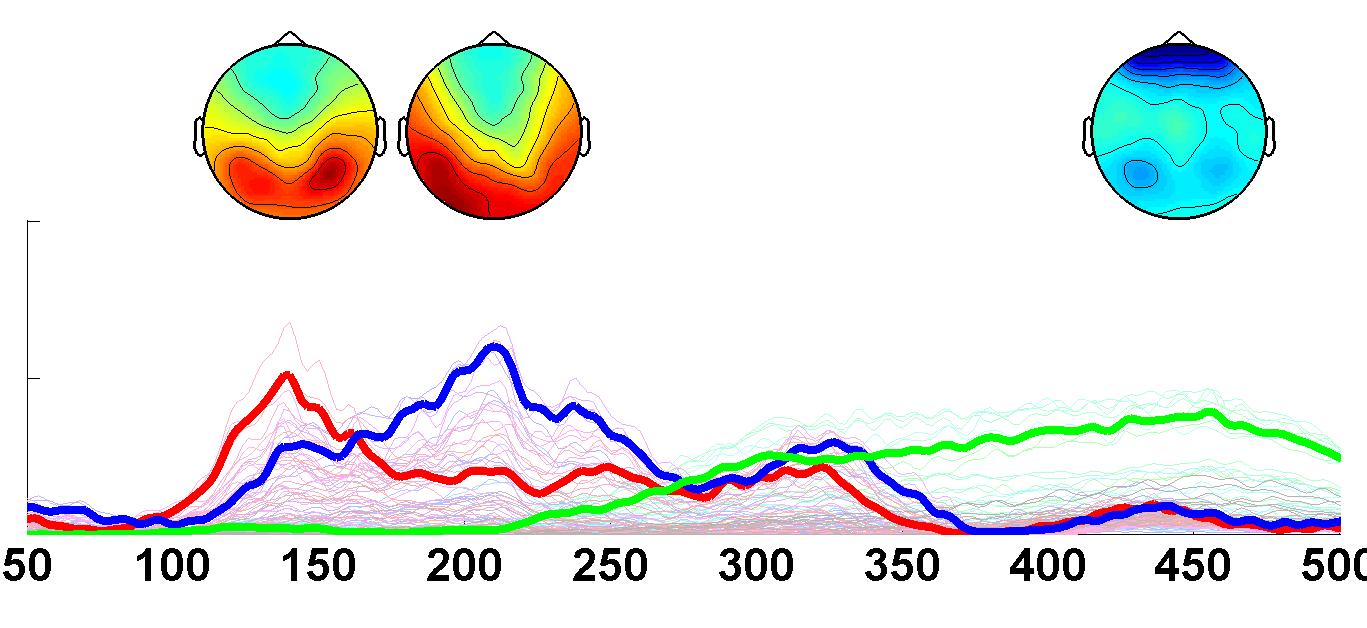}\\

\includegraphics[width=0.3\textwidth, height=20mm]{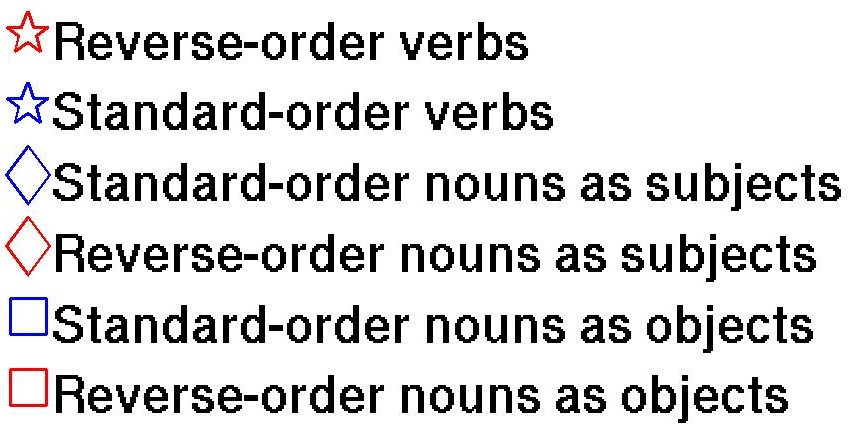}&\\
\end{tabular}
\end{center}
\caption{Results of the FCA: content words (upper), all nouns and standard-order verbs (middle), all verbs (lower). On the left, maps representing the first two axes. On the right, curves of the electrode contributions (in bold: mean curves of the most representative electrodes of the components of the contribution). Located above each maximum values of the contribution is the corresponding topography of the Axis-1 eigenvector, using the same color scale throughout the time window under study (green stands for zero, blue stands for negative values, red stands for positive values).}
\label{fig2}
\end{figure}

The results of the analysis on all 64 conditions pooled were as follows. On Axis 1 (44\% of the inertia), there was a clear-cut distinction between verbs in reverse-order phrases (16 conditions) and all other constituents (48 conditions) (Figure \ref{fig2} upper left). The profile of the dynamic of this distinction has three principal components (Figure \ref{fig2}, upper right): the first (span $80-370 ms$, maximum at $150 ms$) mainly involved the right occipital electrodes; the second (span $100-370 ms$, maximum at $210 ms$) involved the left occipital and temporal electrodes; the third and most important (span $210-550+ ms$, maximum at $460 ms$) involved the prefrontal electrodes and the mid- and left frontal electrodes (over the same span for the latter, but with reversed positivity with respect to the prefrontal electrodes). 

We conducted two supplementary analyses to study the other processes at play in the noun and verb conditions. The first included all units except reverse-order verbs (Figure 2 middle). The first axis of this analysis (20\% of the inertia) distinguished standard-order verbs from all nouns (subjects and objects in standard or reverse order). This distinction spanned two intervals, $100-250 ms$ and $200-400 ms$, and had three maximum values, the highest of which was located at around $250-270 ms$. The main electrodes involved were left and right peripheral electrodes (68, 73, 39, 115, 108, 122, etc.) in the interval $100-250 ms$, and left and right occipito-parietal electrodes (65, 59, 58, 90, 66, 96, 70 etc.) in the interval $200-400 ms$. The topology of the set of coordinates on this axis (Figure \ref{fig2}, middle right) — where on one side we see verbs (second word in standard phrases) and on the other, a tighter cluster of objects (third word in standard phrases) and the remaining noun categories (first and third words in standard and reverse phrases) — shows that this distinction does not reflect word order in the phrase but rather a separate kind of processing that distinguishes the functions of content words in phrases. Moreover, the observed profile of the electrode-time units remained stable when, in addition to standard-order verbs, the analysis included standard-order nouns only or reverse-order nouns only. This suggests that Axis 1 of this FCA reflects differentiation of verbs and nouns.

	The second supplementary analysis (Figure \ref{fig2} lower) deals solely with verbs (standard and reverse orders). The results are similar to that found in the preceding analysis on all words (Figure \ref{fig2} upper), with 40\% of the inertia for this axis. This shows that the greatest processing difference among the content words was a difference between reverse verbs and other content words, regardless of their function in the phrase.
	
	These three analyses revealed the primacy of verb processing: in reverse-order phrases, verbs took precedence over all other words, and in standard-order phrases, verbs took precedence over all nouns, producing distinct profiles in each case. The total span for the distinction between reverse-order verbs and other words was $80-550+ ms$ (so extending past the end of the word), with a smaller total span ($100-400 ms$) for the distinction between standard-order verbs and nouns. The highly distinctive neural processing found here for verbs in reverse-order phrases shows that the underlying process took precedence over the other processes. This can be interpreted non-antinomically either as a process of syntactic movement and argument- (or actant-)schema assignment in reverse-order phrases, as a process triggered by the placement of the relative pronoun before the verb, or as a nesting process also involving calculation of an argument schema. Both of these interpretations involve moves between the syntactic-tree nodes associated with the constituents of the phrase, so we are dealing here with high-level operations within the syntactic processing hierarchy. 
	
	Moreover, the fact that the first distinction between the reverse and standard orders is located at the level of verb processing points back to one of Chomsky's proposals \cite{chomsky1968198}, namely, that the verb controls the assignment of theta roles. Similarly, this fact also points back to Tesnière's \cite{tesniere1965elements}  actantial schema in which the verb acts as the organizing principal of sentences.
	
	The remaining axes obtained in the overall analysis (all words) reflected either the properties of nouns (Section 5.2) or the properties of verbs (Section 5.3).

\subsection{Nouns}
\begin{figure}
     \begin{center}
     \begin{tabular}{  p{5cm}  p{9cm} }
     \hline
\multicolumn{2}{l}{All nouns (axes 1 and 2)}\\

\includegraphics[width=0.3\textwidth, height=20mm]{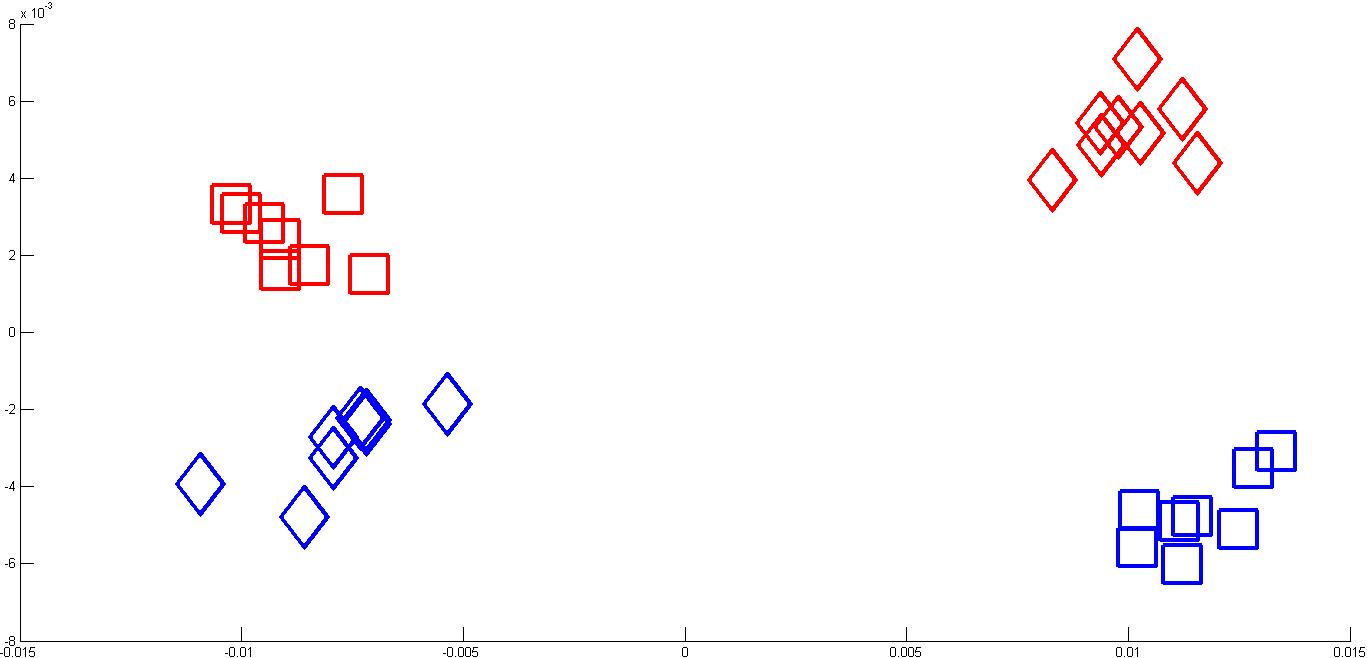}&\includegraphics[width=0.7\textwidth, height=20mm]{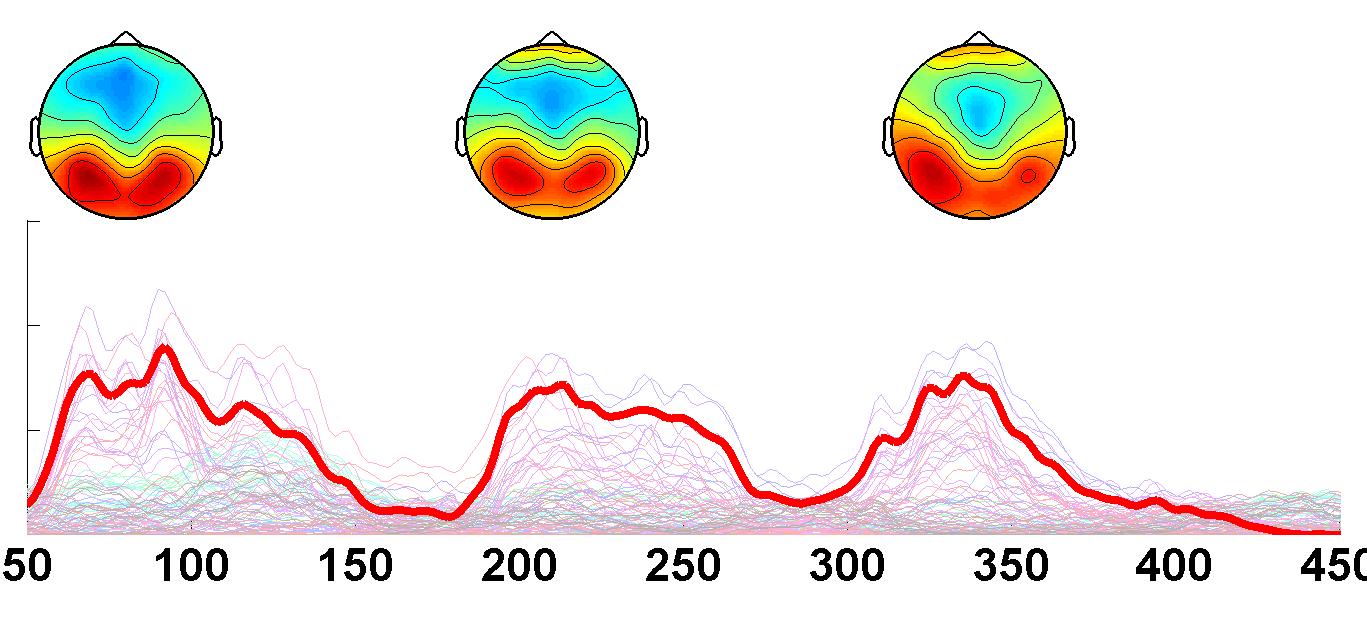}\\
\multicolumn{2}{l}{All nouns (axes 2 and 3)}\\
\includegraphics[width=0.4\textwidth, height=20mm]{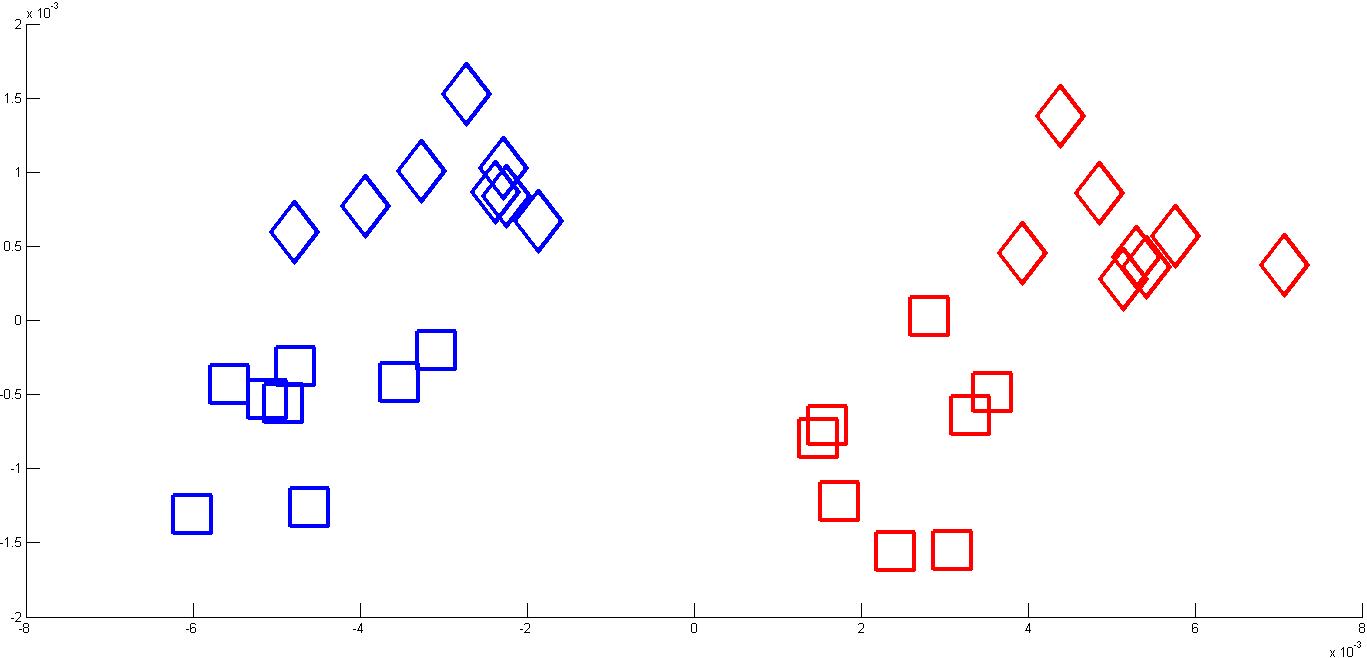}&\includegraphics[width=0.7\textwidth, height=20mm]{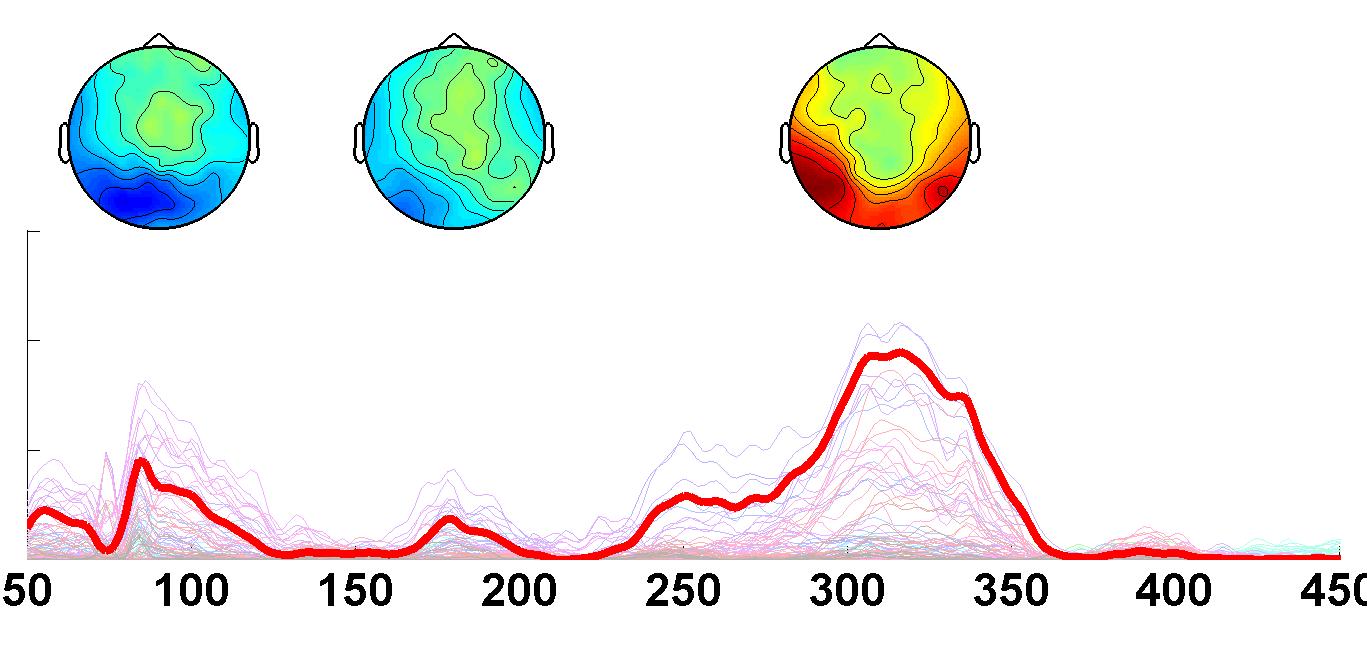}\\
\includegraphics[width=0.3\textwidth, height=20mm]{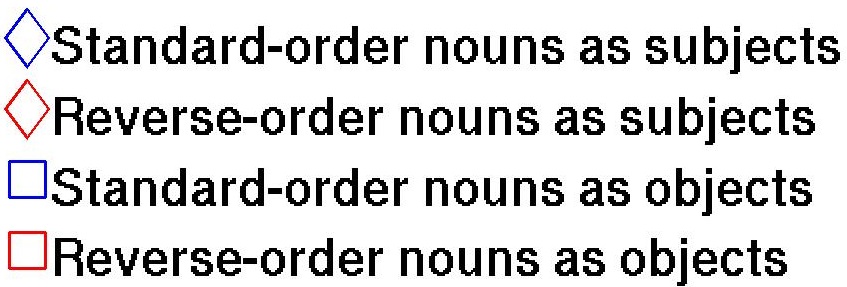}&\includegraphics[width=0.7\textwidth, height=20mm]{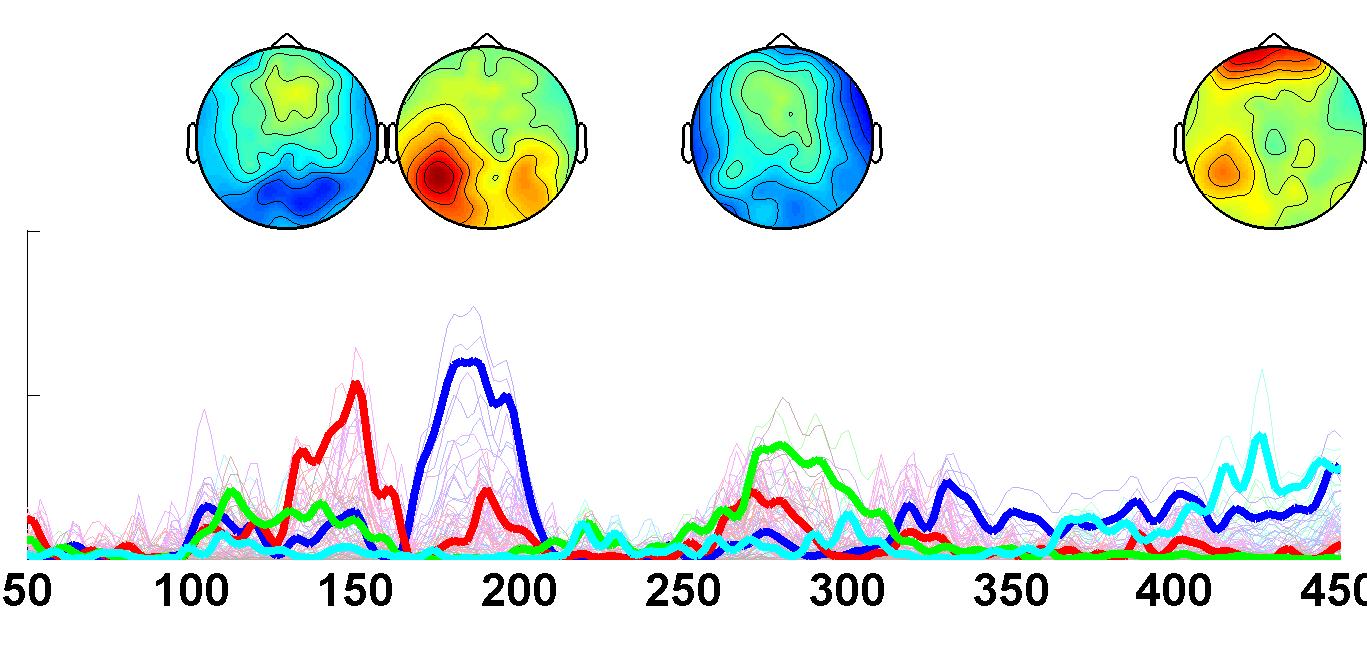}\\
\end{tabular}
\end{center}
\caption{On the left: results of the FCA for all nouns. Coordinates on Axes 1 and 2 (upper); coordinates on Axes 2 and 3 (middle). On the right: curves of the electrode contributions. Axis 1: distinction by position in phrase (upper). Axis 2: distinction between standard order and reverse order (middle). Axis 3: distiction between grammatical subjects and grammatical objects (lower). Above the maximum values of each contribution is the corresponding topography of the eigenvector. }

\label{fig3}
\end{figure}

The analysis of all nouns (Figure \ref{fig3}) revealed the following, in decreasing order of importance (as determined by the rank of the axes): (a) a position-based distinction on Axis 1 (20\% of the total inertia) separating nouns in phrase-initial position from nouns in phrase-final (or third) position; (b) a distinction between standard-order and reverse-order phrases on Axis 2 (13\% of the total inertia) separating all subject or object nouns occurring in standard-order phrases from all nouns occurring in reverse-order phrases); and (c) a distinction between grammatical subjects and grammatical objects on Axis 3 (6\% of the total inertia) separating subjects (first word in standard-order phrases, third word in reverse-order phrases) from objects (third word in standard-order phrases and first word in reverse-order phrases). Axis 1 (separation by position in phrase) spans the interval $50-400 ms$.
 The neurophysiological dynamic of the eigenvector is composed of three intervals, each involving the same occipital electrodes with, however, a tendency toward increasing asymmetry on the left and implication of frontal electrodes also starting with the second interval ($190-270 ms$). Axis 2 (separation of standard and reverse orders) spans $80-370 ms$. The interval that contributes the most is located between 220 ms and 360 ms. The neurophysiological dynamic of the eigenvector essentially involves left temporal electrodes, with a polarity inversion before and after 220 ms. Axis 3 (separation of grammatical subjects and objects) spans 100 ms and its main interval is located $120 ms$ and $210 ms$. The dynamic is composed of four groups of electrodes (shown in red, blue, green, and cyan in Figure \ref{fig3}) (lower). It alternates between two patterns, one composed of right occipital electrodes and also left and right frontal electrodes, the second composed of left temporal electrodes. The two patterns are of opposite polarity. 

	It thus appears that the main processes are ones that pertain first to form and then to syntax. Their dynamics are independent and orthogonal since the eigenvectors of the axes are orthogonal.

\subsubsection{The question of syntactic-semantic interactions and the meaning of nouns}
In the Ploux et al. \cite{ploux2012} study on the processing of isolated words, we were able to draw up a map\footnote{The words were the same as the ones used in the phrases of the present study} whose first axis was person-centered. This axis opposed people (the most distinctive category on this axis), \emph{clothing}, and \emph{parts of the body}, to world entities like \emph{fruits/vegetables} and \emph{tools}, which on this axis were opposed the most to the \emph{people} items. The second axis in the FCA separated biological entities from non-biological ones. Based on the above results (a person-centered reference frame and biological vs. non-biological opposition), which seem to be stable across a diversity of linguistic systems (see the analogous results obtained for Chinese \cite{plouxsubmitted}), we will now present the interactions between syntax and semantics found in the processing of our phrase-embedded nouns. Two calculation methods were chosen. The first was aimed at finding out whether the semantic aspects of the nouns were preserved on the axes, which as we have just seen, represent form and syntax processing. The second method was used to analyze the organization into categories by averaging across the four conditions: (1) grammatical subjects in phrase-initial position (standard), (2) grammatical objects in phrase-final position (standard), (3) grammatical subjects in phrase-final position (reverse), (4) grammatical objects in phrase-initial position (reverse).

\subsubsection{Syntactic-semantic interactions on the first four axes of the noun analysis}
Here, we looked at whether the order of the semantic categories was the same on the first five FCA axes, for all four syntax and position conditions (phrase-initial subjects in standard phrases, phrase-final objects in standard phrases, phrase-initial objects in reverse phrases, phrase-final subjects in reverse phrases). For Axis 1 (position of nouns), the correlation coefficients indicated non-preservation of the semantic structure. For Axis 2 (standard vs. reverse), subjects and objects in standard-order phrases were correlated in all eight semantic categories ($corr(catsubject std,catobj std) = 0.77$, $p < 0.03$ between $110-450 ms$). Also for this axis, there was an off-centered grouping of the parts-of-the-body, clothing, and people categories for nouns in standard-order phrases. This grouping did not exist for nouns in reverse-order phrases. The topology of the people-centered semantic categories in standard-order nouns, together with the coefficient of the correlation between the semantic categories in subject or object position in standard-order phrases, suggest that the reverse-order syntax and/or argument-schema processing interact with person-centered processing. Lastly, on Axis 4, not shown here but on which we found no organization related to form or syntax, we can see a correlation between reverse-order subjects and reverse-order objects: $corr(catsubj. rev.,catobj.rev.) = 0.77$, $p < 0.03$) for all eight semantic categories.

\subsubsection{Analysis by category}
\begin{figure}
     \begin{tabular}{  p{6cm}}
    All categories \\
\includegraphics[width=0.6\textwidth]{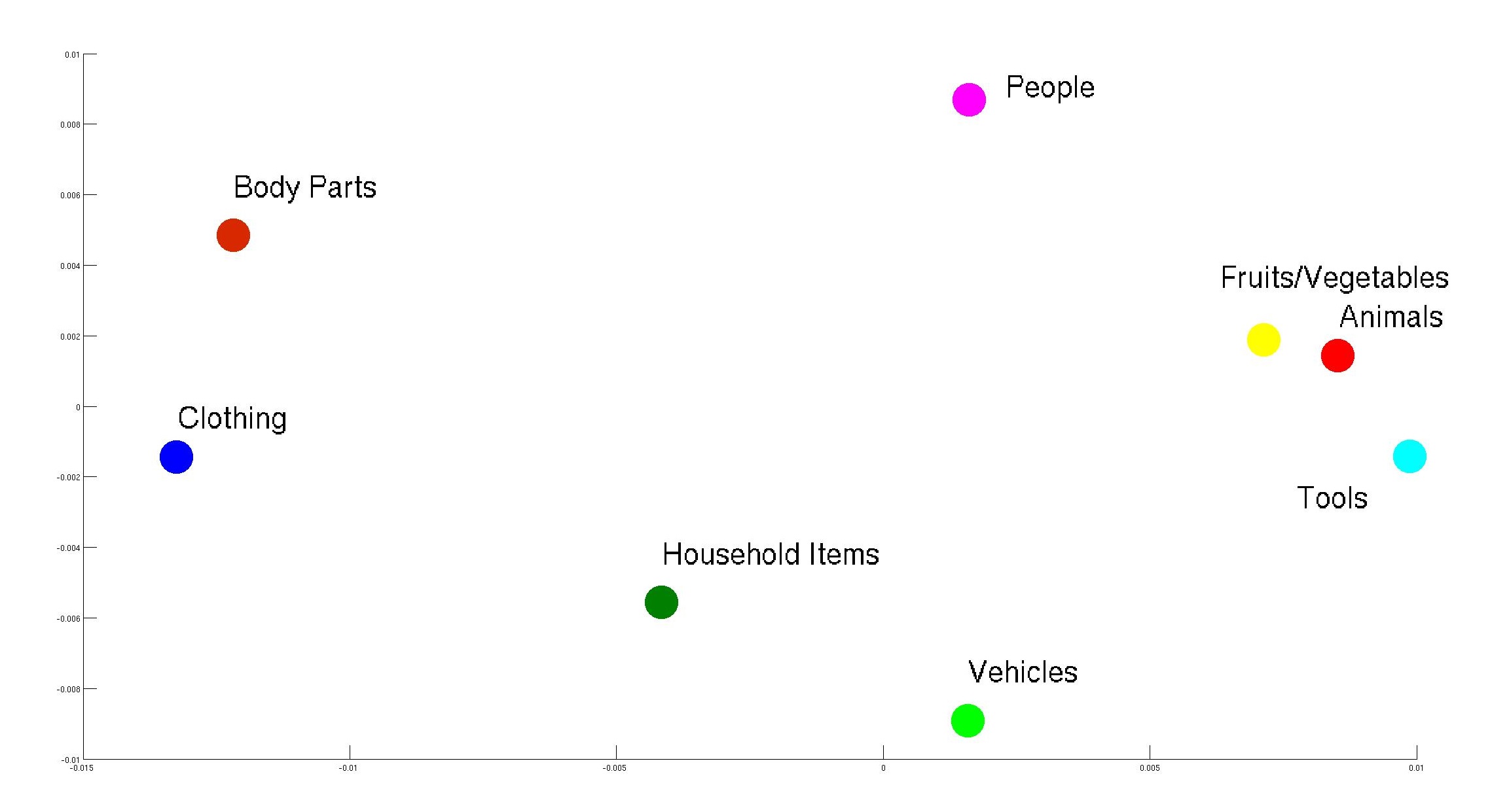}\\
\end{tabular}
\caption{Results for the first two axes of the FCA on the mean of the eight categories ($50-450 ms$). }
\label{fig4}
\end{figure}

We analyzed the eight categories by averaging, for each one, the ERPs in the four conditions: subject-standard, subject-reverse, object-standard, object-reverse. The results (see Figure \ref{fig4}) are highly similar\footnote{However, we can see on the maps that whether the words were isolated or presented in phrases, differences showed up, such as the position of nouns referring to \emph{people} or \emph{animals}. In the map obtained from isolated words, nouns referring to \emph{people} on Axis 1 were the most off-centered in the direction of \emph{parts of the body} and \emph{clothing}, whereas nouns referring to \emph{people} in phrases were located closer to the center of this axis. This could be due to the fact that the mean profile of grammatical subjects (standard or reverse) was very close to the \emph{people} profile, and that the close-to-the-mean profiles of \emph{people} nouns in two of the conditions caused re-centering of this category in the analyses. Likewise, isolated words in the \emph{animals} category were the most off-centered on Axis 2 (biological vs. non-biological), whereas this was no longer the case here. To understand this difference one would have to determine all of the effects of the form and syntax conditions on this category. The third and final difference was that on isolated words, there was less dispersion for the non-biological than for the biological, which was absent here. Once again, the dispersion was greater in the subject conditions (which could reflect the attribution, to non-biological entities, of features related to the ability to be the actant of a verb). In this case, the effect could show up on the mean of the conditions tested.}to those obtained for isolated words \cite{ploux2012}. On both maps, Axis 1 (21\% of the total inertia) opposes clothing and parts of the body to fruits/vegetables and tools, and Axis 2 (here, 16\% of the total inertia) opposes the biological to the non-biological. An Anova on the map projections of all 32 conditions indicated stability of the biological vs. non-biological opposition ($F = 24.6$, $p < 0.001$).

\subsubsection{Comparison of the category analysis and the noun analysis}
In order to provide support or the observed interaction between syntax and semantics based on the coordinates of the conditions on the axes, we used the $cosine$ (whose absolute value, by definition, varies between $0$ and $1$) to calculate a proximity index representing the distance between each of the eigenvectors of the noun-analysis axes to each of the eigenvectors of the category-analysis axes. The values obtained are given in Table \ref{table1}.

\begin{table}
 \begin{tabular}{p{5cm}p{4cm}p{4cm}}
\hline
&$EGV1_{cat}$ (people$-$centered reference frame)&$EGV2_{cat}$ (biological vs. non-biological)\\
$EGV1_{nouns}$ (position)&0.13 &0.19\\
$EGV2_{nouns}$ (std. vs rev)&\emph{0.34}&0.11\\
$EGV3_{nouns}$ (gram. function)&0.05&0.02\\
$EGV4_{nouns}$&$\emph{0.76}$&$0.23$\\
$EGV5_{nouns}$&$0.05$&$\emph{0.34}$\\
\hline

\end{tabular}
\caption{Cosine of the noun-analysis eigenvectors (rows) and the category-analysis eigenvectors (columns). $EGV1$, ..., $EGV5$ stand for eigenvectors 1 to 5, respectively. \emph{cat} stands for category.}
\label{table1}
\end{table}

The FCA noun eigenvectors of Axis 1 (order) and Axis 3 (grammatical subject vs. object) are almost orthogonal ($cos < 0.19$) to the FCA category eigenvectors of Axis 1 (\emph{clothing} and \emph{parts of the body} vs. \emph{fruits/vegetables} and \emph{tools}) and Axis 2 (biological vs. non-biological). This suggests that the processes underlying the semantic-category organization are independant of the processes underlying word order in the phrases or the opposition between grammatical subjects and objects. Note that EGV2std,rev of Axis 2 in the noun FCA (standard vs. reverse) and the EGV1people of Axis 1 in the category FCA (distinguishing \emph{parts of the body} and \emph{clothing} from the other categories) has a $cosine$ of $0.34$. This indicates partial sharing of the electrode-time units involved in these different processes, and supports the results found in the previous section on the correlation between the coordinates on Axis 2 (std vs. rev) for nouns in standard-order phrases and the grouping of three categories (\emph{clothing}, \emph{parts of the body}, and \emph{people}) for subjects and objects in standard-order phrases. In addition, the eigenvector of Axis 4 in the noun analysis and Axis 1 in the category analysis both have a high proximity value ($cos = 0.76$), and the eigenvectors of Axis 5 and Axis 2 (biological vs. non-biological) have a mean proximity of $0.34$.

	To summarize the results, the most important linguistic components of in-phrase noun processing are the ones that discriminated, firstly, the word's position in the phrase; secondly, the structure of the syntactic tree (standard vs. reverse); and thirdly, the word's grammatical function (subject vs. object). Beyond these form- and syntax-based processes, we found traces of semantic proximities on Axis 2 (standard vs. reverse) that were related to the people, clothing, and parts-of-the-body categories and seem to be disrupted for nouns in reverse-order phrases. Lastly, an overall semantic organization showed up when the form and syntax conditions were averaged.
	In short, syntactic processes seem to take precedence and be independent of each other, while interacting with the semantic organization, which they can disrupt. Accordingly, none of the eigenvectors exhibited a strong link with the biological vs. non-biological opposition, which was "diluted" on all axes. Note, however, that $EGV4_{nouns}$ (Axis-4 eigenvector) and $EGV1_{cat}$ (Axis-1 eigenvector) are very close ($cos = 0.76$), and that $EGV2_{cat}$ (Axis-2 eigenvector) has its greatest proximity to $EGV5_{nouns}$ (Axis-5 eigenvector). This shows that even though traces of semantic processing were found on Axes 1 to 3, semantics were essentially secondary to form and syntax (see above).

\subsection{Verbs}
As stated above in Section 5.1, the analysis of all verbs pointed out the important role played by reverse-order verbs. For this reason, we separated the verbs in standard-order phrases from the verbs in reverse-order phrases in order to look at processes other than those induced by this major distinction. An earlier EEG study allowed us to see that verb semantics are difficult to uncover when the parameter used to vary the conditions is not the arguments' semantics but solely the verb's semantics (i.e., the word itself) \cite{Gobin}. A possible explanation is that verb processing may be less distributed in the cortex than noun processing, in such a way that using the EEGs of electrodes spread across a large grid does not provide useful measures for analyzing the semantics of the predicate itself. In our case here, the parameters of the conditions were defined (see Section 5.1) by the category of the verb-dependent arguments. For the analyses conducted on verbs, it is important to note that each condition is the average of all verbs whose first (vs. second) argument pertains to a given category. This implies that the other argument (second or first) can itself pertain to a very broad category, such as being biological or non-biological. We can thus expect an effect of the meaning of the other argument. However, this effect (effect of the category of the verb's other argument in the analysis under consideration) is surely not as strong as the effect of the arguments withe a single, constant category, since such an effect would depend on conditions that vary.

\subsubsection{Standard-order verbs}
Axis 1 (12\% of the total inertia) is difficult to interpret from the semantic point of view\footnote{However, note that for the verbs (standard), there was an opposition between verbs whose subject was a vehicle and verbs whose subject was a household item or a fruit/vegetable. This can be interpreted in terms of the mobile or stationary aspect of the argument, but an additional study is needed to validate this idea.}. On Axis 2 (see Figure \ref{fig5}) (10\% of the total inertia), all verbs with non-biological subjects or objects are located to the left of the origin, whereas verbs pertaining to biological subjects or objects (except for noun objects referring to people or parts of the body) are located to the right. This configuration is stable for the whole set of subjects (sign test, $p < 0.001$) in terms of the differences between the Axis-2 coordinates of verbs whose subject or object is non-biological, and the coordinates on that same axis of verbs whose subject or object is biological (except for objects referring to people or parts of the body). The profile of Axis 2 (Figure \ref{fig5}, upper left) indicates a more scattered neurophysiological process than those obtained above; its primary component is located between $390$ and $420 ms$ (maximum at $405 ms$). The electrodes with the greatest contribution are the left frontal and right occipito-parietal electrodes, along with the mid-frontal electrodes.

\subsubsection{Reverse-order verbs}
\begin{figure}
     \begin{center}
     \begin{tabular}{  p{5cm}  p{9cm} }
     \hline
\multicolumn{2}{l}{All standard-order verbs}\\
\includegraphics[width=0.3\textwidth, height=20mm]{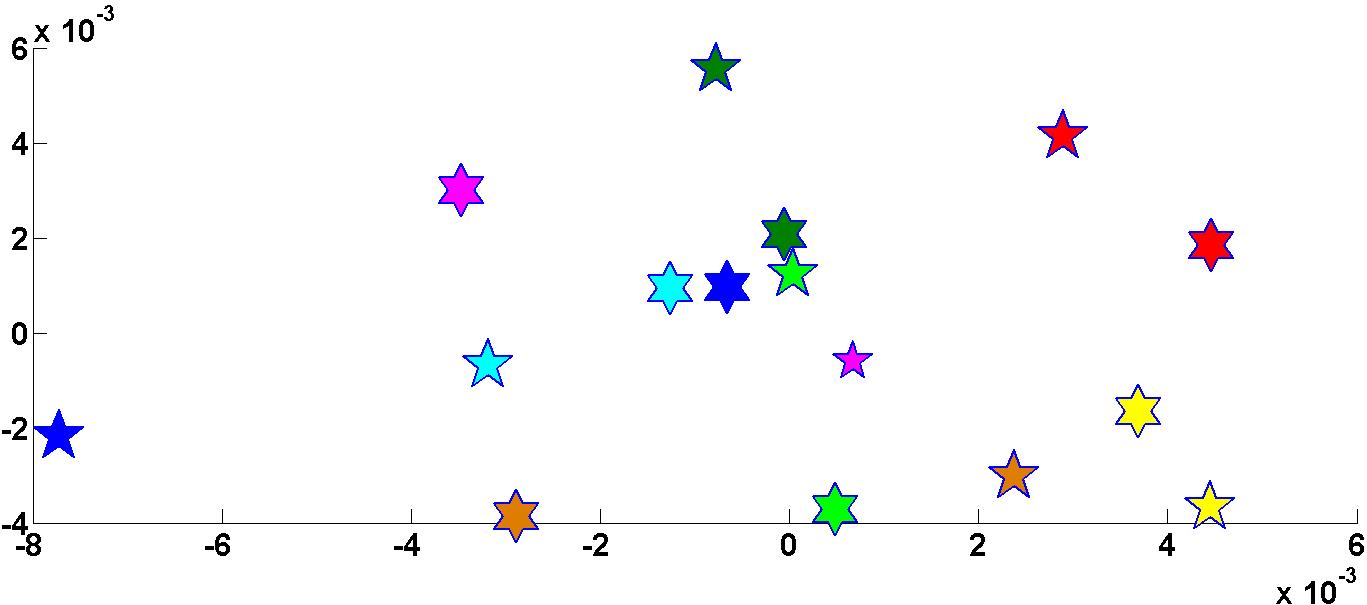}&\includegraphics[width=0.7\textwidth, height=20mm]{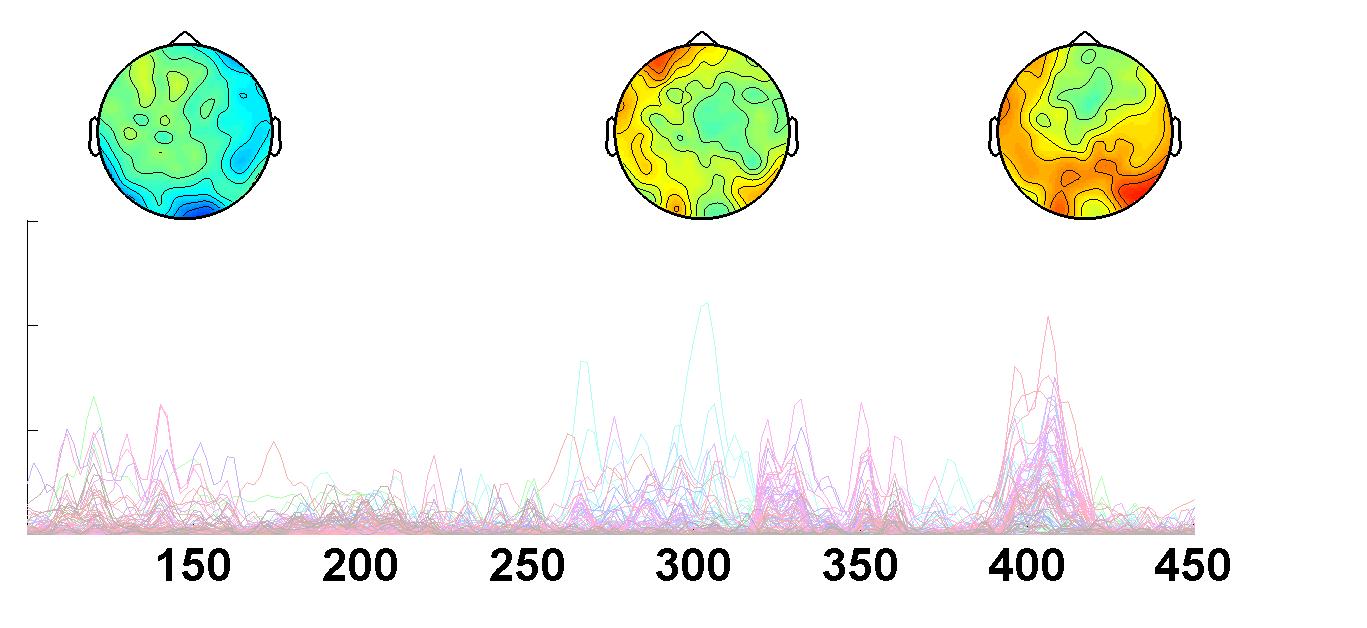}\\

\multicolumn{2}{l}{All reverse-order verbs}\\
\includegraphics[width=0.3\textwidth, height=20mm]{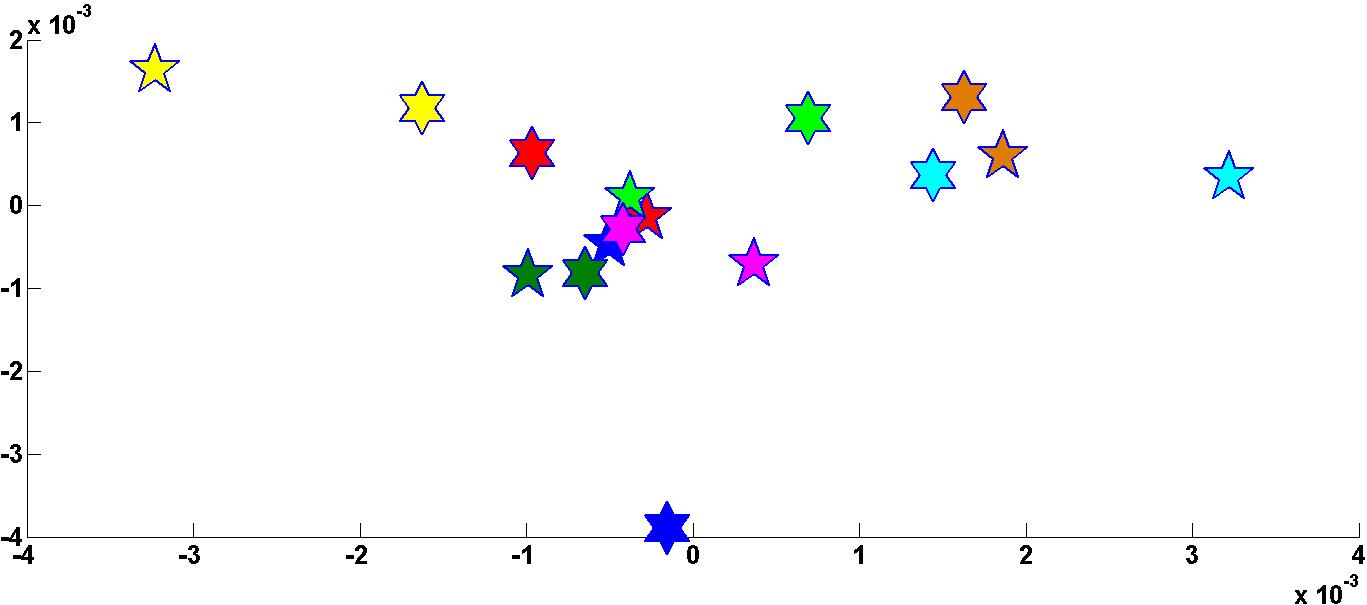}&\includegraphics[width=0.7\textwidth, height=20mm]{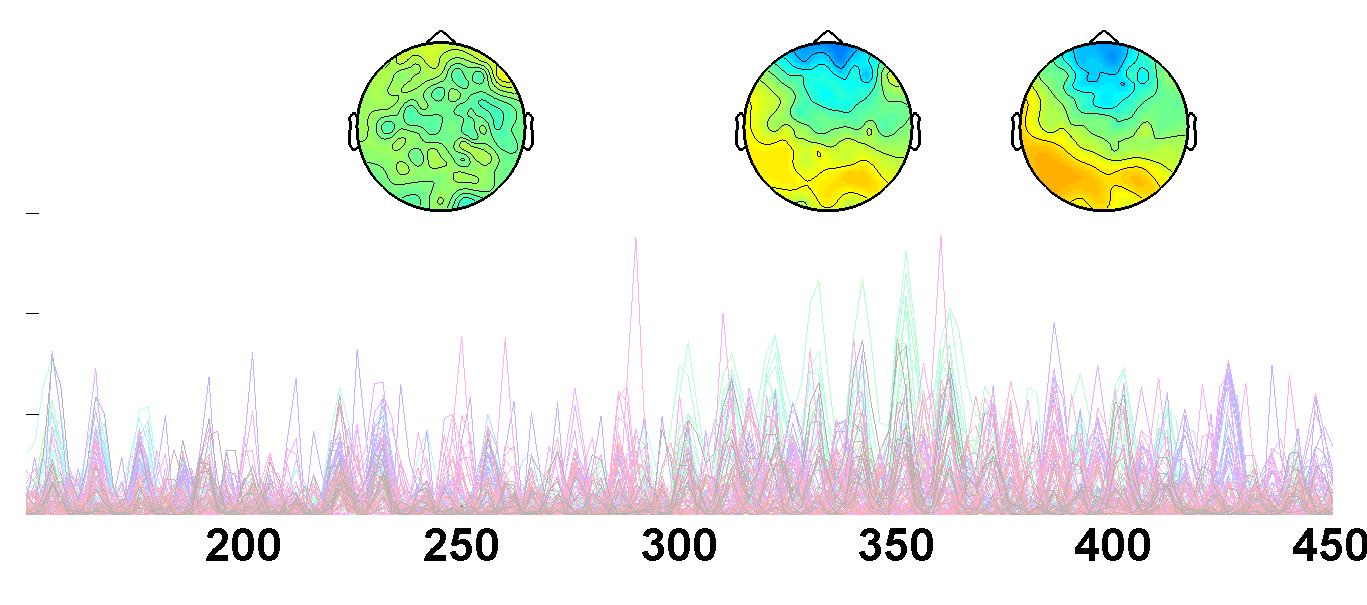}\\
\includegraphics[width=0.3\textwidth, height=20mm]{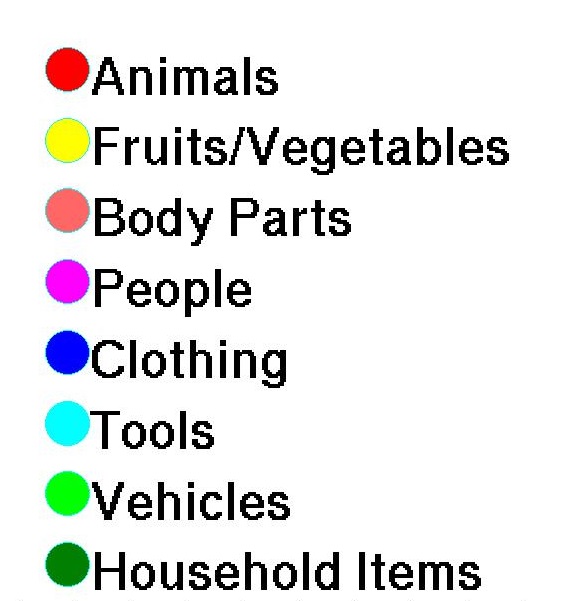}\includegraphics[width=0.3\textwidth, height=20mm]{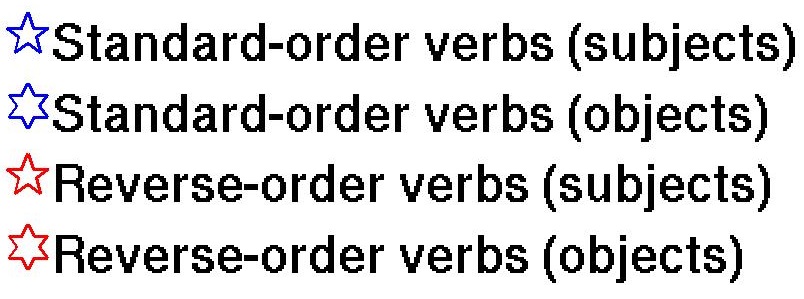}&\\

\end{tabular}
\end{center}
\caption{On the left, results of an FCA (Axes 2 and 3) for standard-order verbs (upper) and reverse-order verbs (lower). On the right, curves of the electrode contributions. Above each one, the corresponding topography of the Axis-2 eigenvector to the maximum values of the contributions.}
\label{fig5}

\end{figure}

As above, Axis 1 (17\% of the total inertia) is difficult to interpret from the semantic viewpoint. Figure 5(wasT5) gives the map obtained from the reverse-order verb FCA and the results for Axis 2 (11\% of the total inertia) and Axis 3 (10.5\% of the total inertia). We did not find a tendency to distinguish the biological from the non-biological. On the other hand, on Axis 2, conditions whose arguments are in the same category are correlated ($coeff = 0.87, p < 0.01$). This shows again that during the $500 ms$ following the display of the verb, the semantic aspects of the arguments are processed. Here again, the profile of Axis 2 (Figure \ref{fig5} lower left) seems more scattered than the profiles involving mainly the phrase's form and syntax. The electrodes with the greatest contribution are the right occipito-parietal ones. 

\subsubsection*{}
	In sum, despite a possible effect that was sometimes the opposite of the effect of the chosen argument's category, we again obtained a semantics-based map 
linking verbs whose subject was a word in the category in question with verbs whose object was a word in that same category. Verb processing thus seems to bring to bear the semantic aspects of the two verb arguments. Note also that this FCA allows us to interrelate the electrode patterns at different times in the neurophysiological dynamic. However, at this stage, we still cannot contend that the observed semantic effect is the result, for a subject or an object, of a process that has the same time course for subjects and objects. It could be, for example, that one and the same semantic pattern that discriminates between biological and non-biological arguments is carried out at a different time within the verb-display window, depending on whether the word is the subject or the object of the phrase. 

	Given that for standard and reverse verbs, we found traces of semantic processing of the verb's arguments, we can suggest that in addition to its role as a phrase organizer, the neurophysiological process associated with verb processing reflects the dynamic \cite{thom2} of the interaction between the arguments, and that this interaction integrates elements of the semantics of the arguments themselves. For example, in order to be understood and processed by the neural system, the dynamic of the interaction between the fillette (little girl) and the pomme (apple) described by the verb manger (to eat) in the phrase \emph{la fillette mange la pomme} vs. \emph{la pomme que mange la fillette} (the apple eaten by the little girl) would have to integrate the specific qualities of both arguments. This finding is also in line with Tesniere's \cite{tesniere1965elements} proposal that the verb expresses a "dramatic event"\footnote{As in a drama play indeed, there is necessarily an event (action and state) and in most cases, actors and circumstances" \cite{tesniere1965elements}.} that mixes semantics and syntax by relating the actants (here, the subject and the object) to each other in a specific manner.

\section{Summary and discussion of neurophysiological processes}
\begin{figure}
\includegraphics[width=10cm]{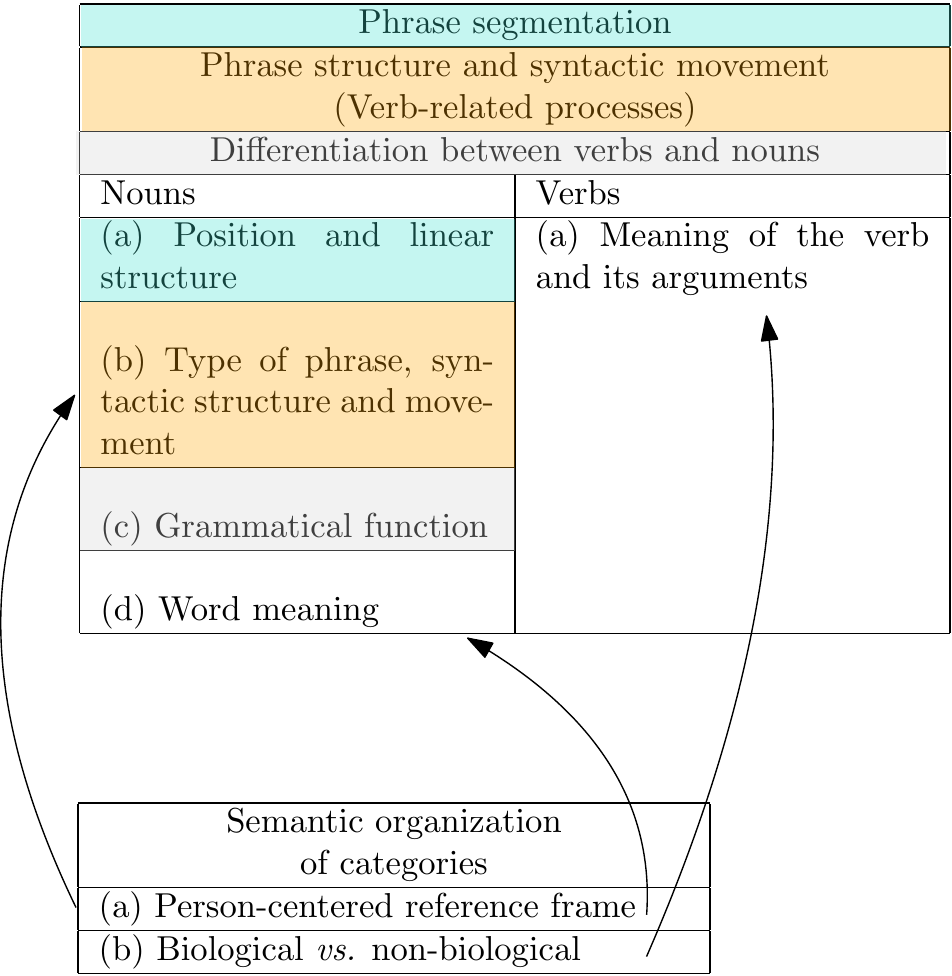}
\caption{Summary of the processes and their interactions}
\label{fig6}
\end{figure}

Different processes with different time spans and distinct electrode patterns were found here for the reading of simple phrases. Our analyses detected these processes, their hierarchy, and their interactions. Figure \ref{fig6} provides a summary of these results.

We can draw the following conclusions about the processes at play:

\begin{itemize}
\item The processing of form (separation into words and position of words in the phrase) seems to take precedence over other types of processing (syntactic and semantic). 
\item	Verbs are at the heart of phrase syntax, namely, the phrase's structure (standard vs. reverse) and the syntax and semantics of the event and its arguments. In addition, the results argue in favor of a common core shared by the verb's argument structure and the verb-based event. 
\item	In the noun analysis, we obtained the same ranking: form first, then syntax (structure of the syntactic tree, followed by grammatical function), and finally, semantic processing, with a possible interaction between the reversing the word order or movement within the phrase, and semantic processes derived from a person-centered reference frame (parts of the body and clothing). We also saw massive orthogonality between the processing of form and syntax on one hand, and between the different syntactic processes on the other. 
\item	Lastly, the observed processing hierarchy translates into neurophysiological dynamics that are increasingly complex and more and more fragmented, not only across time (spans, intervals, and frequencies) but also in terms of their spatial configuration on the scalp (see the topographies of the different eigenvectors). We found greater time spans for processes high up in the syntax tree, and shorter time spans for the grammatical functions of nouns. Also, the semantic processing of nouns and verbs appears to be more fragmented than that of form and syntax.
\end{itemize}

Some of our results are in line with the findings in the neuroscience literature. \cite{fitch2014hierarchical, roy2013syntax} showed that action and corporality share cerebral bases with syntax (more specifically, with syntactic structure processing). \cite{hauk} showed that understanding a word involves activation of the brain areas associated with its semantic content. Taken together, these two findings are consistent with the interaction we detected between the syntactic hierarchy (through variation of the syntactic tree and the notion of movement within that tree) and words with a person-centered reference frame (parts of the body and clothing). Moreover, this interaction did not occur for the grammatical functions (subject and object). Lastly, the fragmentation of semantic processing in time and space aligns with studies (such as \cite{pulvermuller2003,huth2016natural} ) on the cerebral dispersion of the semantic processing of words, and with studies that have found higher frequencies for semantic processing than for syntactic processing \cite{rohm2001role}.

	Similarly, as stated throughout this article, our analyses corroborate the principles and results of studies in linguistics, not only the predominance of syntax as proposed by Chomsky but also — and in some sense contradictory to the idea that syntax and semantics are separate — the role of the verb as the organizing principal of the phrase and the event it describes. Coming back to the debate about modularity, the results of our analyses argue not only in favor of the separation and hierarchization of form and syntax but also of syntax and semantics (except for the above-mentioned special case of the interaction between the processing of syntactic-tree variations and the processing of the person-centered reference frame). However, this separation was not present here when verbs were at stake: the results for verbs indicate that there is a "semantics of syntax" (i.e., the semantic features of the arguments seem to be incorporated into those of the verb). Finally, this study does not allow us to say whether this separation, when present, in the EEG signal measures, is intrinsic or results from differentiation occurring during the development of initially shared structures. 
	
	We are not suggesting that semantics play a minor role in linguistics. The purpose of a word or phrase is to convey a meaning. However, the surface form and the architecture of a phrase's syntactic tree are elements of this conveyor, so they must be processed to gain access to meaning.
	 
	In sum, as stated in the introduction, we have shown that the neurophysiological dynamics and profiles underlying language processing and language structures share characteristics that follow from a structure made up of hierarchically organized components.

\bibliographystyle{apalike}

\begin{thebibliography}{}

\bibitem[Chomsky, 1957]{chomsky1957syntactic}
Chomsky, N. (1957).
\newblock {\em Syntactic structures}.
\newblock Walter de Gruyter.

\bibitem[Chomsky, 1968]{chomsky1968198}
Chomsky, N. (1968).
\newblock 198 1.
\newblock {\em Lectures on government and binding}.

\bibitem[Culioli et~al., 1970]{culioli1970considerations}
Culioli, A., Fuchs, C., and P{\^e}cheux, M. (1970).
\newblock {\em Consid{\'e}rations th{\'e}oriques {\`a} propos du traitement
  formel du langage: tentative d'application au probl{\`e}me des
  d{\'e}terminants, par Antoine Culioli... Catherine Fuchs et Michel
  P{\^e}cheux}.
\newblock Association Jean-Favard pour le d{\'e}veloppement de la linguistique
  quantitative.

\bibitem[Dapretto and Bookheimer, 1999]{dapretto1999form}
Dapretto, M. and Bookheimer, S.~Y. (1999).
\newblock Form and content: dissociating syntax and semantics in sentence
  comprehension.
\newblock {\em Neuron}, 24(2):427--432.

\bibitem[Fitch and Martins, 2014]{fitch2014hierarchical}
Fitch, W. and Martins, M.~D. (2014).
\newblock Hierarchical processing in music, language, and action: Lashley
  revisited.
\newblock {\em Annals of the New York Academy of Sciences}, 1316(1):87--104.

\bibitem[Frisch et~al., 2004]{frisch2004word}
Frisch, S., Hahne, A., and Friederici, A.~D. (2004).
\newblock Word category and verb--argument structure information in the
  dynamics of parsing.
\newblock {\em Cognition}, 91(3):191--219.

\bibitem[Gobin et~al., 2008]{Gobin}
Gobin, A., Paulignan, Y., Cheylus, A., and Ploux, S. (2008).
\newblock A computational method for calculating semantic distances using
  electrophysiological signals.
\newblock {\em xxx}.

\bibitem[Grodzinsky, 2000]{grodzinsky2000neurology}
Grodzinsky, Y. (2000).
\newblock The neurology of syntax: Language use without broca's area.
\newblock {\em Behavioral and brain sciences}, 23(1):1--21.

\bibitem[Hauk et~al., 2004]{hauk}
Hauk, O., Johnsrude, I., and F., P. (2004).
\newblock Somatotopic representation of action words in human motor and
  premotor cortex.
\newblock {\em Neuron}, 39(41):301--307.

\bibitem[Huth et~al., 2016]{huth2016natural}
Huth, A.~G., de~Heer, W.~A., Griffiths, T.~L., Theunissen, F.~E., and Gallant,
  J.~L. (2016).
\newblock Natural speech reveals the semantic maps that tile human cerebral
  cortex.
\newblock {\em Nature}, 532(7600):453--458.

\bibitem[Huth et~al., 2012]{huth2012continuous}
Huth, A.~G., Nishimoto, S., Vu, A.~T., and Gallant, J.~L. (2012).
\newblock A continuous semantic space describes the representation of thousands
  of object and action categories across the human brain.
\newblock {\em Neuron}, 76(6):1210--1224.

\bibitem[Ji and Ploux, 2005]{jibd1}
Ji, H. and Ploux, S. (2005).
\newblock Bases de donn\'ees lexicales contextuelles pour le fran\c{c}ais et
  l'anglais.
\newblock {\em D\'ep\^ot \`a l'APP}.

\bibitem[Kim and Sikos, 2011]{kim2011conflict}
Kim, A. and Sikos, L. (2011).
\newblock Conflict and surrender during sentence processing: An erp study of
  syntax-semantics interaction.
\newblock {\em Brain and Language}, 118(1):15--22.

\bibitem[Kutas et~al., 1980]{kutas1980reading}
Kutas, M., Hillyard, S.~A., et~al. (1980).
\newblock Reading senseless sentences: Brain potentials reflect semantic
  incongruity.
\newblock {\em Science}, 207(4427):203--205.

\bibitem[Malaia and Newman, 2015]{malaia2015neural}
Malaia, E. and Newman, S. (2015).
\newblock Neural bases of syntax--semantics interface processing.
\newblock {\em Cognitive neurodynamics}, 9(3):317--329.

\bibitem[Mitchell et~al., 2008]{Mitchell}
Mitchell, T., Shinkareva, S., Carlson, A., Chang, K.-M., Malave, V., Mason, R.,
  and Just, M. (2008).
\newblock Predicting human brain activity associated with the meanings of
  nouns.
\newblock {\em Science}, 320:1191--1195.

\bibitem[Musso et~al., 2003]{musso2003broca}
Musso, M., Moro, A., Glauche, V., Rijntjes, M., Reichenbach, J., B{\"u}chel,
  C., and Weiller, C. (2003).
\newblock Broca's area and the language instinct.
\newblock {\em Nature neuroscience}, 6(7):774.

\bibitem[Pattamadilok et~al., 2016]{pattamadilok2016role}
Pattamadilok, C., Dehaene, S., and Pallier, C. (2016).
\newblock A role for left inferior frontal and posterior superior temporal
  cortex in extracting a syntactic tree from a sentence.
\newblock {\em cortex}, 75:44--55.

\bibitem[Ploux et~al., 2012]{ploux2012}
Ploux, S., Dabic, S., Paulignan, Y., Cheylus, A., and Nazir, T. (2012).
\newblock {Toward a {Neurolexicology:} A Method for Exploring the Organization
  of the Mental Lexicon by Analyzing Electrophysiological Signals}.
\newblock {\em The {Mental Lexicon}}, 7(2):210--236.

\bibitem[Ploux et~al., 2016]{plouxsubmitted}
Ploux, S., Wang, R., Zhong, Z., Cheylus, A., Zhao, H., Xin, Y., and Lu, B.-L.
  (2016).
\newblock Structural stability of lexical semantic spaces: Nouns in chinese and
  french.
\newblock {\em submitted}.

\bibitem[Pulverm{\"u}ller, 2003]{pulvermuller2003}
Pulverm{\"u}ller, F. (2003).
\newblock {\em {The Neuroscience of Language:} On Brain Circuits of Words and
  Serial Order}.
\newblock Cambridge University Press.

\bibitem[Pulverm{\"uller} et~al., 2005]{Pulvermuller2005}
Pulverm{\"uller}, F., Shtyrov, Y., and Ilmoniemi, R. (2005).
\newblock {Brain signatures of meaning access in action word recognition:} an
  {MEG} study using the mismatch negativity.
\newblock {\em Journal of Cognitive Neuroscience}, 17(6):884--892.

\bibitem[Rastier, 2005]{Rastier2005}
Rastier, F. (2005).
\newblock Mésosémantique et syntaxe.
\newblock {\em Revue {\'e}lectronique Texto}.

\bibitem[R{\"o}hm et~al., 2001]{rohm2001role}
R{\"o}hm, D., Klimesch, W., Haider, H., and Doppelmayr, M. (2001).
\newblock The role of theta and alpha oscillations for language comprehension
  in the human electroencephalogram.
\newblock {\em Neuroscience letters}, 310(2):137--140.

\bibitem[Roy et~al., 2013]{roy2013syntax}
Roy, A.~C., Curie, A., Nazir, T., Paulignan, Y., Des~Portes, V., Fourneret, P.,
  and Deprez, V. (2013).
\newblock Syntax at hand: common syntactic structures for actions and language.
\newblock {\em PloS one}, 8(8):e72677.

\bibitem[Tesni{\`e}re, 1965]{tesniere1965elements}
Tesni{\`e}re, L. (1965).
\newblock El{\'e}ments de syntaxe structurale.

\bibitem[Thom, 1980]{thom2}
Thom, R. (1980).
\newblock {\em Mod\`eles math\'ematiques de la morphog\'en\`ese}.
\newblock Christian Bourgeois Editeur, Paris.

\bibitem[Tyler et~al., 2011]{tyler2011left}
Tyler, L.~K., Marslen-Wilson, W.~D., Randall, B., Wright, P., Devereux, B.~J.,
  Zhuang, J., Papoutsi, M., and Stamatakis, E.~A. (2011).
\newblock Left inferior frontal cortex and syntax: function, structure and
  behaviour in patients with left hemisphere damage.
\newblock {\em Brain}, 134(2):415--431.

\bibitem[van Dam and Desai, 2016]{van2016semantics}
van Dam, W.~O. and Desai, R.~H. (2016).
\newblock The semantics of syntax: The grounding of transitive and intransitive
  constructions.
\newblock {\em Journal of cognitive neuroscience}.

\bibitem[Wehbe et~al., 2014]{wehbe2014simultaneously}
Wehbe, L., Murphy, B., Talukdar, P., Fyshe, A., Ramdas, A., and Mitchell, T.
  (2014).
\newblock Simultaneously uncovering the patterns of brain regions involved in
  different story reading subprocesses.
\newblock {\em PlosOne}.

\bibitem[Zaccarella et~al., 2017]{zaccarella2017building}
Zaccarella, E., Meyer, L., Makuuchi, M., and Friederici, A.~D. (2017).
\newblock Building by syntax: the neural basis of minimal linguistic
  structures.
\newblock {\em Cerebral Cortex}, 27(1):411--421.

\end{thebibliography}

\end{document}